\pdfoutput=1

\documentclass[11pt]{article}

\usepackage[preprint]{acl}

\usepackage{times}
\usepackage{latexsym}

\usepackage[T1]{fontenc}

\usepackage[utf8]{inputenc}

\usepackage{microtype}

\usepackage{inconsolata}

\usepackage{amsmath,amsfonts,bm}

\def\eqref#1{equation~\ref{#1}}

\def\1{\bm{1}}

\DeclareMathAlphabet{\mathsfit}{\encodingdefault}{\sfdefault}{m}{sl}
\SetMathAlphabet{\mathsfit}{bold}{\encodingdefault}{\sfdefault}{bx}{n}

\DeclareMathOperator*{\argmax}{arg\,max}

\usepackage{hyperref}
\usepackage[ruled]{algorithm2e}
\usepackage{enumitem}
\usepackage{graphicx}
\usepackage{epsfig}
\usepackage{amsmath}
\usepackage{amssymb}
\usepackage{multirow}
\usepackage{booktabs}
\usepackage{arydshln}
\usepackage{amsfonts}
\usepackage{algpseudocode}
\usepackage{xcolor}
\newcommand{\hclip}{\textsc{Helip}}

\newcommand{\ie}{\textit{i.e.}}

\usepackage{color, colortbl}
\definecolor{LightCyan}{rgb}{0.88,1,1}	
\definecolor{Gray}{gray}{0.95}
\usepackage[symbol]{footmisc}

\newcommand{\tmlrnew}[1]{{\textcolor{black}{#1}}}

\title{Getting More Juice Out of Your Data: Hard Pair Refinement Enhances Visual-Language Models Without Extra Data}

\author{Haonan Wang\textsuperscript{\rm1}\thanks{Equally contributed to this work, $^\dagger$Corresponding author.\protect\\$^\S$Affiliated with Department of Systems Engineering and Engineering Management, and Shun Hing Institute of Advanced Engineering, The Chinese University of Hong Kong, Hong Kong}, Minbin Huang\textsuperscript{\rm2}$^{*\S}$, Runhui Huang\textsuperscript{\rm3}, Lanqing Hong\textsuperscript{\rm4}\textsuperscript{$\dagger$}, Hang Xu\textsuperscript{\rm4}, \\
\textbf{Tianyang Hu\textsuperscript{\rm1},}
\textbf{Xiaodan Liang\textsuperscript{\rm3},} \textbf{Zhenguo Li\textsuperscript{\rm3},} \textbf{Hong Cheng\textsuperscript{\rm2}$^{\S}$,} \textbf{Kenji Kawaguchi\textsuperscript{\rm1}}\\
\vspace{-5pt}\\
\normalsize $^{1}$National University of Singapore \quad $^{2}$The Chinese University of Hong Kong  \\
 \quad $^{3}$Sun Yat-sen University \quad $^{4}$Huawei Noah's Ark Lab \\
}

\begin{document}
\maketitle

\begin{abstract}
\label{abs}
Contrastive Language-Image Pre-training (CLIP) has become the standard for cross-modal image-text representation learning. Improving CLIP typically requires additional data and retraining with new loss functions, but these demands raise resource and time costs, limiting practical use.
In this work, we introduce \textbf{\hclip}, a cost-effective strategy that improves CLIP models by exploiting challenging text-image pairs within existing datasets in continuous training. This eliminates the need for additional data or extensive retraining.
Moreover, \hclip\ integrates effortlessly into current training pipelines with minimal code modifications, allowing for quick and seamless implementation.
On comprehensive benchmarks, \hclip\ consistently boosts existing models. 
In particular, within just two epochs of training, it improves zero-shot classification accuracy on ImageNet for SLIP models pre-trained on CC3M, CC12M, and YFCC15M datasets by $3.05\%$, $4.47\%$, and $10.1\%$ , respectively.
In addition, on fine-grained classification datasets, \hclip\ improves the zero-shot performance of CLIP and SLIP by an average of $8.4\%$ and $18.6\%$, and their linear probe performance by an average of $9.5\%$ and $3.0\%$.
The code is publicly available at: \url{https://github.com/haonan3/HELIP-NAACL-2025.git}.
\end{abstract}


\section{Introduction}
\label{intro}
Contrastive Language-Image Pretraining (CLIP)~\citep{RadfordKHRGASAM21} is quickly becoming the standard for foundation models~\citep{awais2023foundational} due to its effectiveness for a variety of vision-language tasks without task-specific finetuning~\citep{LiSGJXH21, Baldrati_2022_CVPR}.
However, web-crawled image-text pairs used for the CLIP model pretraining are often loosely connected, resulting in multiple plausible matches beyond the assigned ones~\citep{WuCZGGV22}. 
Several methods have been presented to improve CLIP models by investigating appropriate matches and utilizing widespread supervision among image-text pairs for training~\citep{0001LXH22, LiSGJXH21, MuK0X22, DiHT}. 

Efforts to improve contrastive language-image pretraining models have primarily taken two directions: (1) the addition of objectives to improve the efficacy of supervision~\citep{0001LXH22, MuK0X22}; and (2) the employment of intra- and inter-modality similarity to select and retrain models using data deemed challenging at the sample level~\citep{LiSGJXH21, DiHT}. 
However, those approaches inevitably require retraining, and those identified as challenging data struggle to bring benefits to model performance. 
This challenge is partly due to their reliance on finding challenging data within a single batch during training, where truly beneficial challenging data is rare. And, CLIP models' original contrastive loss is not optimally configured to exploit the nuances of difficult data. 
These limitations restrict the practical application of these methods, especially considering the substantial investments already made in pretraining numerous CLIP models~\citep{0001LXH22, MuK0X22}; retraining for minimal gains is inefficient.
This aspect underscores the need for efficient enhancement strategies that do not rely on additional data collection to improve existing pretrained models.


To improve the existing CLIP models, we introduce the \hclip\ framework, which involves further training the models with challenging data selected from their original training dataset.
\hclip\ defines and identifies the challenging data at the pair level, distinguishing it from traditional methods that compare sample-level similarities between images and texts. Specifically, \hclip\ treats each text-image pair as a distinct entity within the joint vision-language space, and defines pairs in close proximity as hard pairs.
Furthermore, \hclip\ introduces the \textbf{Hard Pair Mining (HPM)} strategy, a novel approach that moves beyond the traditional use of representation spaces learned by CLIP models. 
Note, the CLIP space is primarily designed for evaluating sample-level similarities—for instance, comparing an image and text (individually, not as a pair)—lacking in evaluating characteristics at the pair level.
HPM transforms the task of discovering pairs in close proximity into a solvable proxy task, with the goal of selecting a pair set that optimally supports the target pair's text-image agreement.
\hclip\ enhances CLIP models not just with the original text-image contrastive loss~\citep{RadfordKHRGASAM21}, which uniformly pushes all negative samples away from their positive counterpart but also incorporates the \textbf{Hard Negative Margin Loss (HNML)} into the loss function. 
As depicted in Figure~\ref{fig:margin_loss}, HNML imposes an additional geometric structure on the representation space, reflecting the pair-level similarity. Through this approach, \hclip\ effectively leverages the information within challenging data to boost model performance.

Empirical evidence shows that \hclip\ improves the performance of existing CLIP models, including pre-trained CLIP, SLIP, and DECLIP, across a variety of benchmarks, such as zero-shot classification, text-image retrieval, and fine-grained linear probing. 
For zero-shot classification on ImageNet, CIFAR-10, and CIFAR-100, \hclip\ consistently boosts the performance of all six pre-trained models. Particularly, using \hclip\ to boost SLIP models pre-trained on CC3M, CC12M, and YFCC15M results in ImageNet zero-shot accuracy gains of $3.05\%$, $4.47\%$, and $10.14\%$, respectively. Further, on seven fine-grained image classification datasets, those pre-trained models achieve better zero-shot and linear probe performance with \hclip. Specifically, the average zero-shot accuracy of CC3M pre-trained CLIP and SLIP are improved by $8.4\%$ and $18.6\%$. The average linear probe accuracy of CC3M pre-trained CLIP and SLIP are improved by $9.5\%$ and $3.0\%$ respectively. Additionally, the performance gain is also valid in terms of zero-shot retrieval, with $1.1$ of R@1 on Flickr30K, and $2.2$ of R@1 on COCO for SLIP with \hclip.

\section{Related Work}
\label{sec:Related Work}
\textbf{Vision-Language Pre-training.} Vision-Language Pretraining (VLP) leverages large-scale image-text datasets to learn joint representations transferable to downstream tasks. VLP models are typically classified into single-stream and dual-stream architectures. Single-stream models concatenate text and visual features processed by a single transformer~\citep{li2019visualbert, chen2022improving, zhang2020devlbert}. Dual-stream models use separate encoders for image and text, performing cross-modal interactions at a higher level~\citep{RadfordKHRGASAM21, JiaYXCPPLSLD21, LiLZCOSYY22, MuK0X22, YaoHHLNXLLJX22}.
CLIP~\citep{RadfordKHRGASAM21}, a dual-stream model, employs contrastive learning with 400M web-crawled image-text pairs to achieve remarkable zero-shot recognition performance. Recent works enhance CLIP's performance by applying self-supervision within the visual modality~\citep{MuK0X22} or incorporating nearest neighbor supervision~\citep{LiLZCOSYY22}. While these methods improve performance, they introduce additional computational costs due to data augmentations.\\
\textbf{Contrastive Learning with Hard Negative Samples.} Contrastive learning aims to learn representations by bringing similar examples closer and pushing dissimilar ones apart~\citep{chen2020simple, mocov2, wang2020understanding}. Incorporating hard negative samples into the loss function has been shown to improve performance~\citep{negativesnotequal, HuynhKWMK22, KalantidisSPWL20, LiSGJXH21, DiHT, RobinsonCSJ21, ShahSCC22}.
In language-image contrastive learning, approaches like \citet{LiSGJXH21} and \citet{DiHT} mine hard negatives using intra- or inter-modality similarity, selecting samples with high cosine similarity in visual or textual features. However, due to the loose alignment in web-crawled data, high similarity in these features doesn't necessarily indicate that pairs are difficult to distinguish.
In contrast, we propose a hard sample mining method that discovers similar pairs in the joint vision-language space, efficiently selecting truly challenging samples to improve learning.
\vspace{-1pt}
\section{Hard Pairs for Visual-Language Models}
\label{sec:Method}
\vspace{-5pt}
In this section, we first define the notations and revisit CLIP for zero-shot recognition in the preliminary section. Next, we introduce the Hard Pairs Mining strategy (HPM), and the associated Hard Negative Margin Loss (HNML), designed to efficiently exploit hard pairs.

\vspace{-2pt}
\subsection{Preliminaries}
\label{sec:prelim}
\vspace{-2pt}
We consider the task of contrastive image-text pretraining. Given an image-caption dataset $\mathcal{D} = \{ z_i \}_{i=1}^N = \{ (x_i^{I}, x_i^{T}) \}_{i=1}^N $, $(x_i^{I}, x_i^{T}) \in \mathcal{I}\times\mathcal{T}$, the $x_i^{I}$, $x_i^{T}$ denote the image and its corresponding caption, $\mathcal{I}$ and $\mathcal{T}$ indicates visual and textual space respectively,
and $\mathcal{I}\times\mathcal{T}$ indicates the joint Vision-Language space. The goal is to learn a dual encoder model $\phi = \{ \phi_{image}, \phi_{text} \}$, where $\phi_{image}$ represents the image encoder and $\phi_{text}$ denotes the text encoder. We use the shorthand $I_i = \phi_{image}(x_i^{I})$ and $T_i = \phi_{text}(x_i^{T})$ to denote the encoded representation of an image and its caption, respectively.
The contrastive objective of CLIP is formulated as,
\begin{equation}
\begin{small}
\label{equ:clip_loss}
\begin{aligned}
\ell_{\text{CLIP}} = -\frac{1}{|B|} \sum_{i \in B} \log \frac{\exp \left(sim(I_i, T_i) / \sigma\right)}{\sum_{j \in B} \exp \left(sim(I_i, T_j) / \sigma\right)},
\end{aligned}
\end{small}
\end{equation}
where $sim(\cdot,\cdot)$ is the cosine similarity function, $B$ is a batch of samples and $\sigma$ is a trainable parameter controlling the temperature.
Intuitively, the above formulation explicitly aligns the representations of image and text from one pair.

\begin{figure}[t]
\begin{center}
\includegraphics[width=\linewidth]{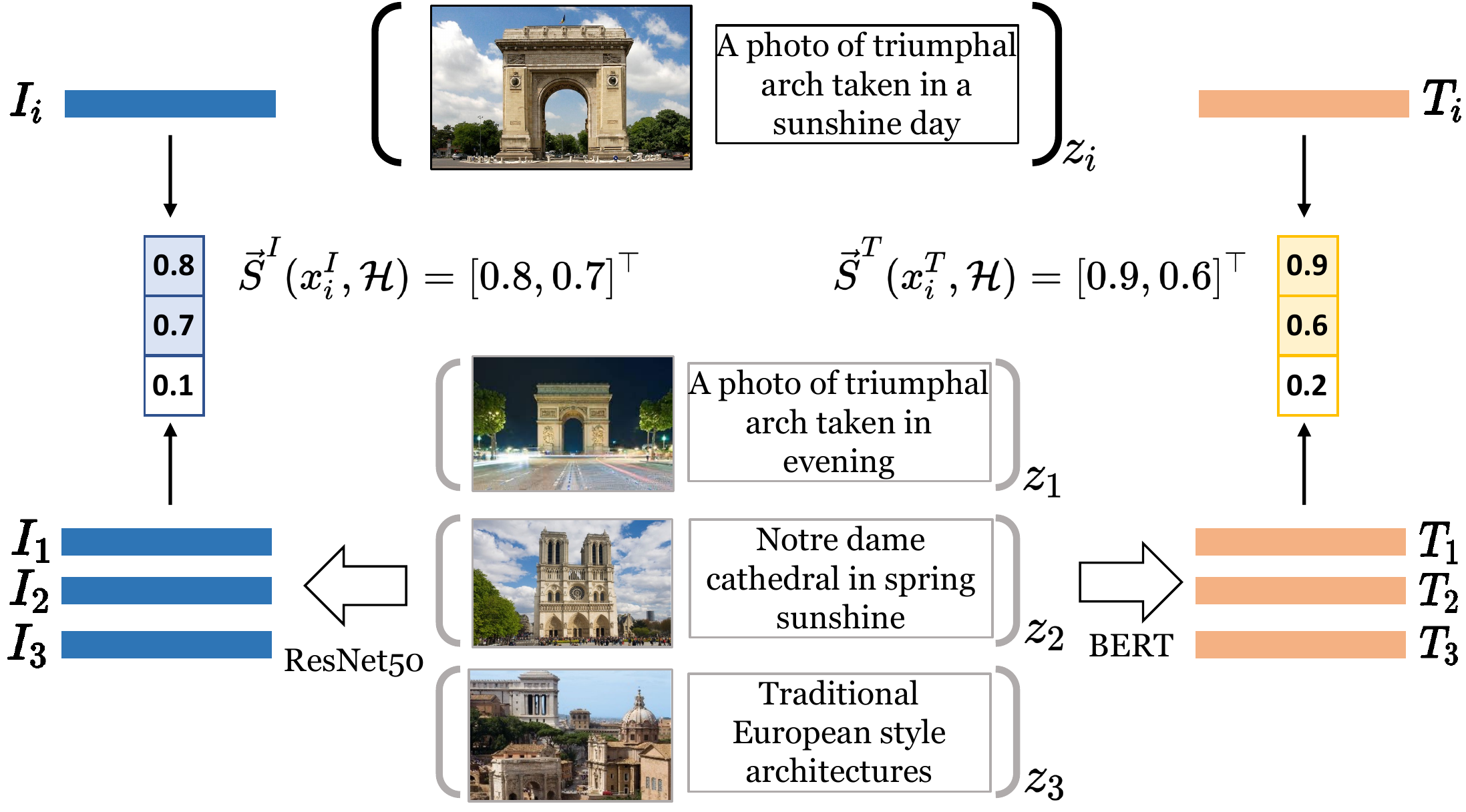}
\vspace{-10pt}
\caption{
{Hard Pair Mining (HPM)}. 
Choose hard pairs by optimizing the support set to maximize the agreement prediction of the target pair.\label{fig:sample_selection}
}
\end{center}
\vspace{-15pt}
\end{figure}

\subsection{HPM: Hard Pair Mining}
\label{sec:hard_sample_mining}
In this study, we define \underline{\textit{hard pairs}} as the pairs that are nearby to a specified target pair within the joint vision-language space, $\mathcal{I}\times\mathcal{T}$, which serves as the domain for pair data. 
Equation~\ref{equ:orig_goal} depicts the problem of hard pair mining. Here, ${z_i}$ represents the target pair, $\mathcal{H}_i$ denotes a set of pairs chosen from the dataset ${\mathcal{D}}_i = \mathcal{D} \setminus {z_i}$, and the metric $\mathbf{S}(,)$ quantifies the similarity between the target pair and a set of pairs,
\begin{equation}
\vspace{-7pt}
\begin{aligned}
\mathcal{H}^{\star}_i  = \argmax_{\mathcal{H}_i} \mathbf{S}(z_i,\mathcal{H}_i).
\end{aligned}
\vspace{-5pt}
\label{equ:orig_goal}
\end{equation}
However, a key challenge arises in defining the similarity metric for pairs, $\mathbf{S}$. 
Existing CLIP methods~\citep{RadfordKHRGASAM21, LiLZCOSYY22, 0001LXH22} preliminary focus on aligning an image with its caption~\citep{RadfordKHRGASAM21, 0001LXH22} from a image-text pair. They rarely emphasize on bringing similar pairs closer while distancing the dissimilar ones, which makes current methods fall short in gauging similarity between two pairs. For instance, the cosine similarity between two pairs is ill-defined, within the context of current methods.

To identify nearby pairs, we introduce the idea of text-image pair agreement maximization. This can be viewed as a proxy task for selecting hard pairs. 
To illustrate the rationale for using text-image pair agreement as a proxy for selecting hard pairs, we return to the principle obtained from traditional machine learning methods: the prediction of a model on a test sample is substantially influenced by samples in the training dataset that are similar to the test one. 
For example, the K-Nearest Neighbors (KNN) algorithm classifies a new instance using the K-closest training examples. The linear regression model predicts the output of a test sample using the weighted sum of the training samples, with higher weights given to samples that are more similar to the test sample. 
Recent empirical and theoretical studies on model memorization and generalization~\citep{chen2009similarity, zhang2021understanding, stephenson2021geometry, brown2021memorization} also provide support for this.
Intuitively, if a pair agreement prediction model trained on a set of pairs predicts a specific target pair as having a high probability of being a matching pair, the target pair is likely to be similar to the matching pairs on which the model was trained.
The challenge of selecting hard pairs is transformed into an optimization task centered on the text-image pair agreement, which is formally represented as:
\begin{equation}
\vspace{-5pt}
\label{equ:transfer_goal}
\begin{small}
\begin{aligned}
\argmax_{\mathcal{H}_i} \mathbf{S}(z_i,\mathcal{H}_i) = \argmax_{\mathcal{H}_i} P_{\mathcal{M}}(z_i|\mathcal{H}_i),
\end{aligned}
\end{small}
\vspace{-2pt}
\end{equation}
where $P_{\mathcal{M}}(z_i|\mathcal{H}_i)$ denotes the prediction of a pair agreement model,  $\mathcal{M}$, for the pair $z_i$ based on a pair set $\mathcal{H}_i$. This set is a subset of ${\mathcal{D}}_i$.
In this framework, the goal of selecting a hard pair is transformed into identifying a training set $\mathcal{H}_i$ such that the model $\mathcal{M}$ predicts the target pair as a matching pair.


Designing a suitable pair agreement prediction model for this proxy task is a nontrivial endeavor because the model needs to not only predict the pair matching probability but also allow the optimization of the training set, as indicated in Equation~\ref{equ:transfer_goal}. Consequently, a conventional deep neural network design becomes unviable due to the impracticality of retraining across all possible sets $\mathcal{H}_i$ from ${\mathcal{D}}_i$.
Taking inspiration from recent work~\citep{norelli2022asif}, we propose a data-centric design for the agreement prediction model $\mathcal{M}$. As illustrated in Figure~\ref{fig:sample_selection}, the model leverages two pretrained single-modal encoders, i.e., $f_{\text{image}}$ and $f_{\text{text}}$, to align representations of images and texts in a unified Vision-Language space. Specifically, the model encodes the target pair $z_i$ into $(I_i, T_i)$ using these single-modal encoders.
For the visual modality, we determine a similarity vector between the target pair $z_i$ and the dataset ${\mathcal{D}}_i$. The similarity vector is defined as $\vec{S}^{I}(x^I_i, {\mathcal{D}}_i) = [\dots, sim(I_i, I_j), \dots]^{\top} \in \mathbb{R}^{N-1}$. Here $I_j = f_{\text{image}}(x^I_j)$ with $(x^I_j, x^T_j)$ being an element of ${\mathcal{D}}_i$, and function $sim(\cdot, \cdot)$ denotes the cosine similarity. 
To counteract noise, values in the vector $\vec{S}^{I}(x^I_i, {\mathcal{D}}_i)$ are set to zero if $sim(I_i, I_j) < \tau$. This cleaned-up vector is represented as $\widetilde{S}^{I}$. The procedure for the textual modality is analogous, producing a vector denoted as $\widetilde{S}^{T}$. 
Note, the representations in this shared space are intuitively interpretable: each dimension corresponding to the visual/textual similarity of the input to a unique pair in the multimodal dataset. This interpretable characteristic enables us to directly optimize the supporting set to maximize the pair matching probability:
\begin{equation}
\vspace{-1pt}
\label{equ:sample_selection}
\begin{aligned}
\mathcal{H}^{\star}_i = \argmax_{|\mathcal{H}_i|=k} \widetilde{S}^{I}(x^I_i, \mathcal{H}_i)^{\top} \widetilde{S}^{T}(x^T_i, \mathcal{H}_i),
\end{aligned}
\vspace{-1pt}
\end{equation}
where the $\mathcal{H}^{\star}_i$ is the hard pair set and $k \in \mathbb{R}^+$ is the number of selected pairs which is much less than $|\mathcal{D}|$. The previous problem can be efficiently solved by greedily choosing dimensions that maximize the inner product. Due to the interpretable property, the selected dimensions are corresponding to the desired pairs.

\noindent\textbf{Mitigation of Noisy Data Impact.} The prior method assumes the target pair $z_i$ to be a suitable matching pair. However, in inherently noisy datasets, such as web-crawled ones like LAION~\citep{schuhmann2022laion}, mismatched pairs might be present. The potential negative effects of hard pairs generated by these mismatched pairs necessitate the development of a strategy for identifying and eliminating them.
We create a pair removal strategy based on the availability of hard pairs: A target pair $z_i$ is deemed as unsuitable and thus removed, if there is a non-empty subset of the mined hard pair set,  $\mathcal{H}^{sub}_i \subseteq \mathcal{H}^{\star}_i$ with $|\mathcal{H}^{sub}_i| > 0$, such that $\widetilde{S}^{I}(x^I_i, \mathcal{H}^{sub}_i)^{\top} \widetilde{S}^{T}(x^T_i, \mathcal{H}^{sub}_i) = 0$.

Intuitively, this equation suggests that the number of entries positively supporting the target pair $z_i$ as a matching pair is fewer than $k$. 
To illustrate how this concept can aid in cleaning noisy data, consider the following example: Suppose the target pair consists of a ``cat'' image but a ``dog'' caption (clearly it is a mismatch). For it to be considered a correct match, numerous pairings with same erroneous pattern (i.e., ``cat'' images paired with ``dog'' captions) would need to exist in the dataset. By assuming a certain error types are fewer than $k$ throughout the dataset, if no subset of size $k$ within the dataset $\mathcal{D} \setminus z_i$ supports $z_i$ as a matching pair, this signals that the target pair is an outlier, likely due to a labeling error or mismatch.
Such outliers can degrade dataset quality, so they are removed to ensure the reliability of hard data.

\noindent\textbf{Fast Hard Pair Mining (FastHPM).} 
It is intuitive to infer that for a dataset collected from a single source, the number of intrinsic hard pairs, which are robust enough to enhance the learned representation, will proportionally increase with the size of the dataset originating from that source. To identify $k$ (much less than $|\mathcal{D}|$) qualified hard pairs, a portion of the dataset $\mathcal{D}$ is sufficient.
As a result, we present the Fast Hard Pair Mining (FastHPM) approach, which was designed to avoid the time complexity associated with hard pair mining over the entire dataset. 
FastHPM's objective can be formalized as follows:
\begin{equation}
\label{equ:sample_selection_approx}
\begin{aligned}
\mathcal{H}^{\star}_i \approx \argmax_{|\mathcal{H}|=k} \widetilde{S}^{I}(x^I_i, \mathcal{H}_i)^{\top} \widetilde{S}^{T}(x^T_i, \mathcal{H}_i),
\end{aligned}
\vspace{-5pt}
\end{equation}
where $\mathcal{H}_i \subseteq \overline{\mathcal{D}}_i$ and $|\overline{\mathcal{D}}_i|=C$ is sampled uniformly from set ${\mathcal{D}}_i$.
In this equation, it's noteworthy that the selection of value $C$ is solely based on the number of hard pairs $k$, instead of the size of ${\mathcal{D}}_i$. Consequently, this optimization reduces the time complexity of FastHPM to $\mathcal{O}(N)$. 
The detailed procedure of the hard pair mining algorithm is presented in Appendix~\ref{apd:algs}.

\subsection{HNML: Hard Negative Margin Loss}
\label{sec:negative_margin_loss}
\begin{figure}[t]
\centering
\includegraphics[width=\linewidth]{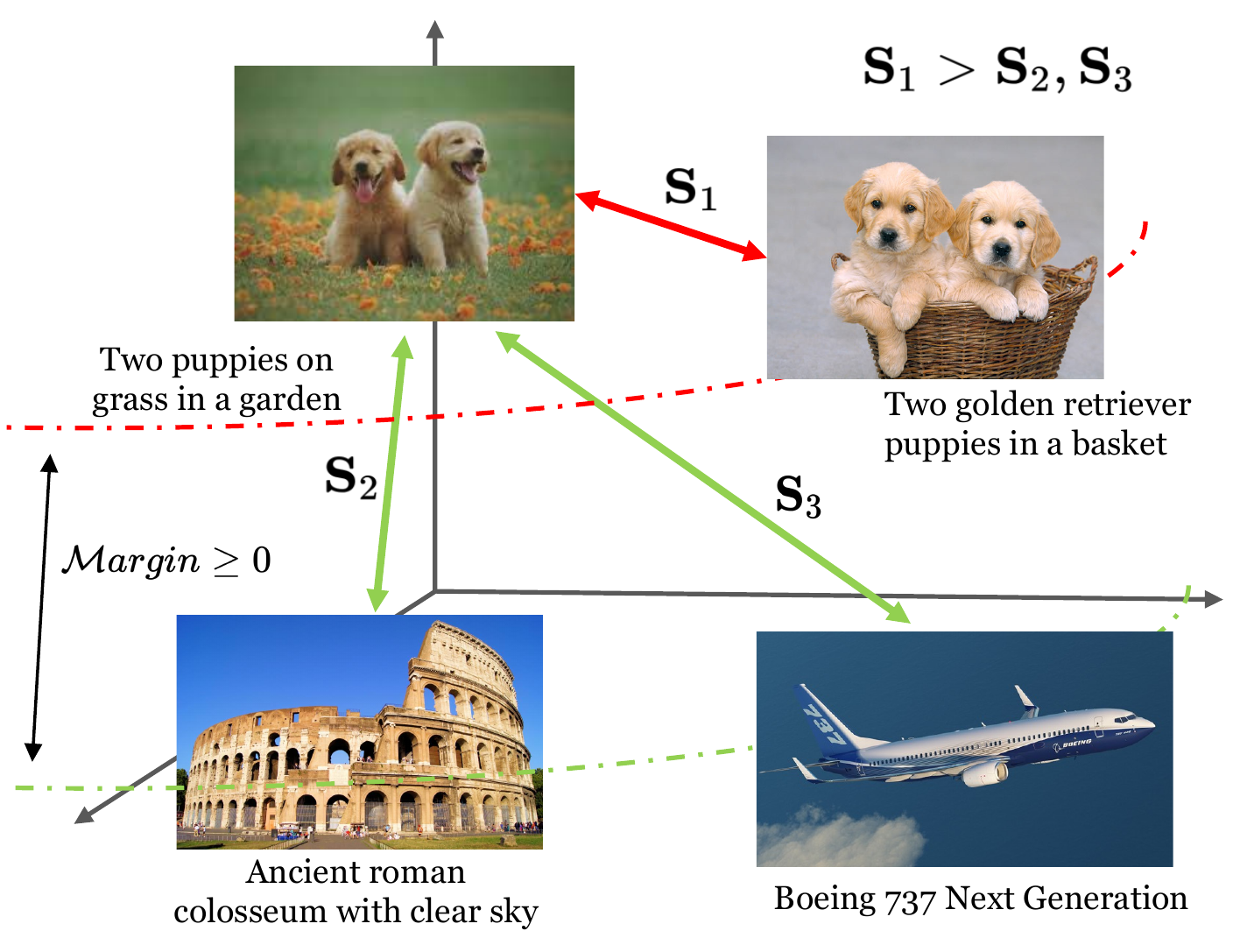}
\vspace{-10pt}
\caption{{Hard Negative Margin Loss (HNML).} Hard negative pairs (\ie, the golden retriever) are closer to the positive than the normal negative pairs.}
\vspace{-15pt}
\label{fig:margin_loss}
\end{figure}
\begin{figure}[t]
\centering
\includegraphics[width=\linewidth]{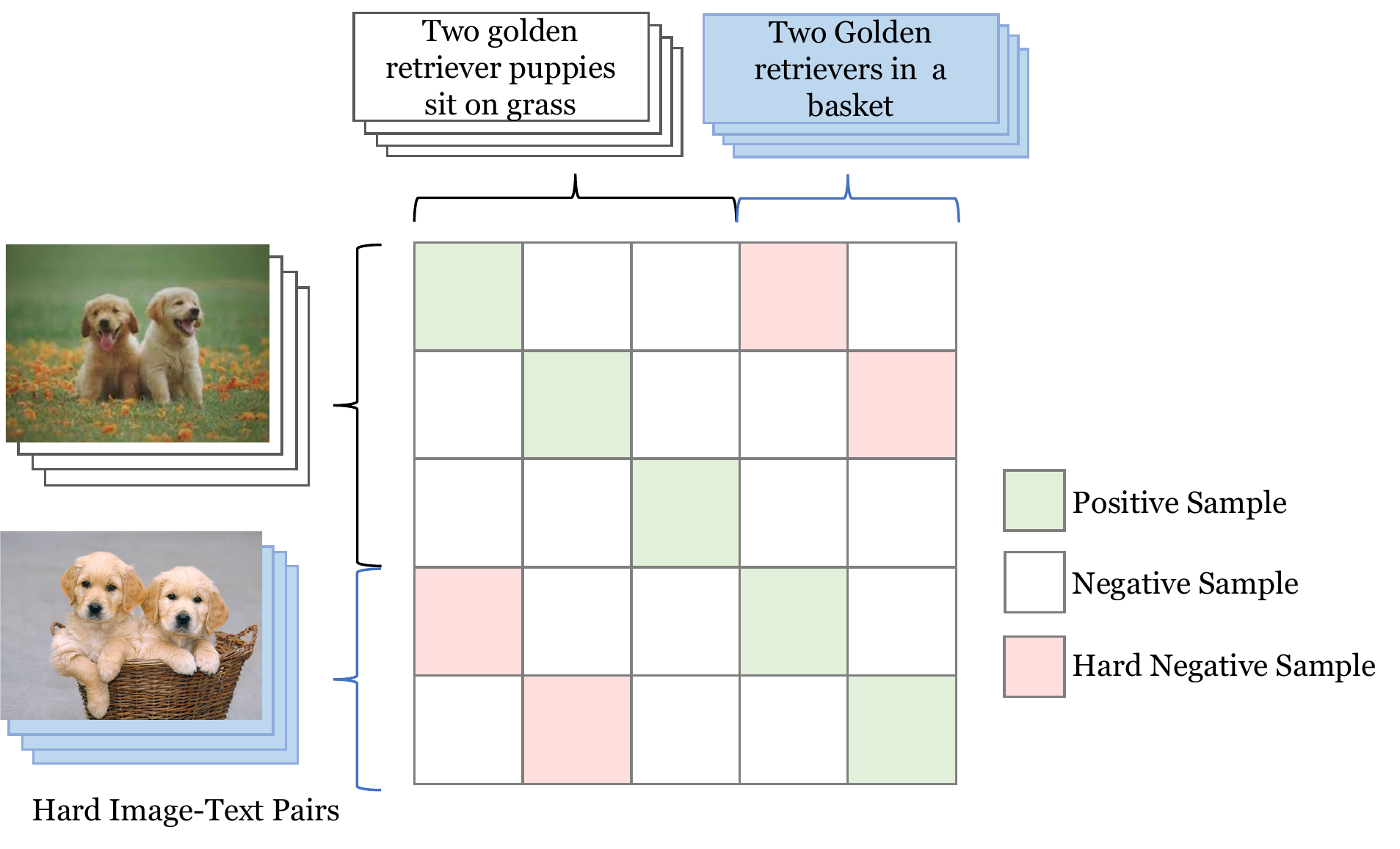}
\vspace{-15pt}
\caption{{Continuous training CLIP with Hard Pairs.} For text-image pairs within a batch, we sample corresponding hard data from the preprocess hard pair set.}
\vspace{-15pt}
\label{fig:cl_vis}
\end{figure}

The image-text contrastive loss $\ell_{CLIP}$, as illustrated in the preliminary section, aligns the true image-text pairs. But it poses no constraints on the overall geometry among data pairs~\citep{goel2022cyclip}.
After involving hard data into the finetuning stage, equally maximizing the distance for normal negative pairs and hard negative pairs is an undesired way to utilize the information provided by hard negative pairs.
The intuition follows directly from Figure~\ref{fig:margin_loss}. In a desired representation space, the similarity between the positive and the hard negative, $\textbf{S}_1$, should be greater than the similarity between the positive and those normal negatives, $\textbf{S}_2, \textbf{S}_3$.
Therefore, to impose the additional geometric structure, we introduce the Hard Negative Margin Loss (HNML):
\begin{equation}
\begin{scriptsize}
\begin{aligned}
\ell_{\text{margin}} =  \frac{1}{|B|} \sum_{j \in B} \max  \big(0, sim(I_i, T_j) - \min_{j' \in \mathcal{H}^p_i } \{sim(I_i, T_{j'})\} \big),
\end{aligned}
\end{scriptsize}
\end{equation}
where $\mathcal{H}^p_i \subseteq \mathcal{H}^{\star}_i$ is the hard negative pairs for the target $z_i$ involved in one training batch.
Note, the HNML is computationally efficient. No extra inner product computation is required. The geometric regularization is applied over the inner product matrix computed in the original CLIP loss, Equation ~\eqref{equ:clip_loss}.
Then, the well-trained model is finetuned with the following loss, where $\gamma$ is the hyperparameter balancing the two losses,
\begin{equation}
\vspace{-6pt}
\begin{aligned}
\ell_{\text{finetune}} =\ell_{\text{CLIP}}  + \gamma \ell_{\text{margin}}.
\end{aligned}
\vspace{-2pt}
\end{equation}
To boost the performance of well-trained CLIP models without introducing extra data and extra parameters, we introduce the further training strategy which involves the preprocessed hard pairs into the batch composition during training.
As shown in Figure~\ref{fig:cl_vis}, for text-image pairs within the batch ${B}$, we randomly sample a subset ${B}'$ as seeds. Then, for 
$z_i \in {B}'$, we randomly select $|\mathcal{H}_i^p| = p$ pairs from $\mathcal{H}^{\star}_i$. The actual training batch is $\overline{B} = {B} \bigcup\limits_{i=0}^{|{B}'|} \mathcal{H}_i^p$. 
We summarize the training pipeline in appendix~\ref{apd:algs}.

\section{Experiments}
\label{sec:exp}
\vspace{-3pt}
In Section~\ref{subsec:main_exp}, we empirically investigate \hclip's efficacy in improving zero-shot classification, image-text retrieval, and linear probing performance of existing vision-language models. Section~\ref{subsec:scaled_exp} examines \hclip's performance with scaled training data. We provide in-depth studies on Hard Positive Mining (HPM) and Hard Negative Mining with Margin Loss (HNML) in Sections~\ref{subsec:margin_loss} and \ref{subsec:hpm_detail}, respectively. Discussions on robustness over noisy datasets and additional empirical analyses of hard mining methods are deferred to the appendix.

\subsection{Experimental Setup}

\textbf{Training Datasets.} We used open-source datasets including CC3M~\citep{SoricutDSG18}, CC12M~\citep{changpinyo2021conceptual}, and two 15M subsets of YFCC100M: v1~\citep{RadfordKHRGASAM21} and v2~\citep{LiLZCOSYY22}. The combined datasets—CC3M, CC12M, and YFCC15M v1 (Open29M~\citep{LiLZCOSYY22})—were not fully obtained due to expired URLs. Additionally, we sampled 7.5M and 8M subsets from the noisier LAION-5B~\citep{schuhmann2022laion}, labeled LAION7.5M and LAION8M. Although smaller than the 400M pairs used in CLIP's original study~\citep{RadfordKHRGASAM21}, these datasets suit our computational resources and have been widely used in language-image pretraining benchmarks~\citep{goel2022cyclip, LiLZCOSYY22, MuK0X22}.

\noindent\textbf{Downstream Datasets.} We evaluate \hclip\ using zero-shot image classification, linear probing, and zero-shot image-text retrieval. Beyond ImageNet~\citep{DBLP:conf/cvpr/DengDSLL009}, CIFAR10, and CIFAR100~\citep{krizhevsky2009learning}, we assess performance on seven fine-grained classification datasets: Caltech101~\citep{fei2004learning}, Food101~\citep{bossard2014food}, Sun397~\citep{xiao2010sun}, Flowers102~\citep{nilsback2008automated}, CUB~\citep{wah2011caltech}, Stanford Cars~\citep{krause20133d}, and FGVC Aircraft~\cite{maji13fine-grained}. For zero-shot image-text retrieval, we use MS-COCO~\citep{DBLP:conf/eccv/LinMBHPRDZ14} and Flickr30K~\citep{DBLP:conf/iccv/PlummerWCCHL15}. Implementation details are provided in the appendix.

\subsection{Main Results and Discussion}
\label{subsec:main_exp}
\noindent \textbf{Zero-Shot Classification.} 
We evaluate the zero-shot performance of CLIP, SLIP, and DECLIP models, along with their \hclip-fine-tuned versions (CLIP-\hclip, SLIP-\hclip, and DECLIP-\hclip), on the CC3M, CC12M, YFCC15M, and Open29M datasets. Table~\ref{tab:main_table} shows that models with \hclip\ consistently outperform their counterparts.
Specifically, on the CC3M dataset, \hclip\ boosts the ImageNet zero-shot accuracy of CLIP from 19.04\% to 19.86\% and improves SLIP by over 13\% to 26.05\%. We include two baseline methods, CYCLIP~\citep{goel2022cyclip} and CLOOB~\citep{CLOOB}, for reference.
Using SLIP checkpoints from~\citet{MuK0X22} pretrained on CC12M, SLIP-\hclip\ achieves a 4.47\% higher zero-shot accuracy on ImageNet than SLIP. Since DECLIP parameters for CC3M and CC12M are unavailable, we compare DECLIP and DECLIP-\hclip\ on the YFCC15M v2 dataset, also presenting SLIP and DECLIP models pretrained by~\citet{LiLZCOSYY22} evaluated with their pipeline (denoted with ${*}$).
Because templates significantly impact zero-shot tasks, we use our evaluation pipeline (the same as OpenCLIP) for fair comparison; further baseline details are in Appendix~\ref{apd:baseline}.
Both SLIP and DECLIP show improvements with \hclip, averaging increases of 15.49\% and 6.74\%, respectively. To demonstrate \hclip's efficacy on larger datasets, we evaluated CLIP and CLIP-\hclip\ on Open29M. The original CLIP model reaches its best zero-shot ImageNet accuracy of 42.32\% at epoch 18. Applying \hclip\ boosts this to 46.33\% with just \textit{one additional epoch}, whereas continued training with the original CLIP loss slightly decreases accuracy to 42.25\%.
\begin{table}[t]
\centering
\scalebox{0.73}{
\setlength{\tabcolsep}{1pt}{
\begin{tabular}{ccccc}
\toprule
\multicolumn{1}{l}{}                          &     Method        & \multicolumn{1}{l}{ ImageNet} & CIFAR10 & CIFAR100 \\ 
\toprule
\multicolumn{1}{c}{\multirow{6}{*}{\rotatebox{90}{CC3M}}}    & CYCLIP~\citep{goel2022cyclip}      & 22.08   & {51.45} & {23.15}         \\
\multicolumn{1}{c}{}                         & CLOOB~\citep{CLOOB}         & 23.97       &    -     &    -                       \\
\multicolumn{1}{c}{}                         & CLIP\tmlrnew{$^{\dagger}$}~\citep{RadfordKHRGASAM21}          & 19.04       & {33.06} & {13.77}                   \\
\multicolumn{1}{c}{}                  \cellcolor{LightCyan}      & CLIP\tmlrnew{$^{\dagger}$}-\hclip   & \cellcolor{LightCyan}19.86       &\cellcolor{LightCyan} {34.05} & \cellcolor{LightCyan} {14.13}                   \\
\multicolumn{1}{c}{}                         & SLIP~\citep{MuK0X22}          & 23.00          & {65.61} & {34.69}                   \\
\multicolumn{1}{c}{}                    \cellcolor{LightCyan}     & SLIP-\hclip  & \cellcolor{LightCyan}\textbf{26.05}       & \cellcolor{LightCyan}\textbf{{68.18} }& \cellcolor{LightCyan}\textbf{{37.77}}                   \\
\midrule
\multicolumn{1}{c}{\multirow{4}{*}{\rotatebox{90}{CC12M}}}   & CLIP\tmlrnew{$^{\dagger}$}~\citep{RadfordKHRGASAM21}        & 30.27   & {51.07} & {21.94}         \\
\multicolumn{1}{c}{}              \cellcolor{LightCyan}        & CLIP\tmlrnew{$^{\dagger}$}-\hclip   & \cellcolor{LightCyan}32.05       & \cellcolor{LightCyan}{52.27} & \cellcolor{LightCyan}{24.51}                   \\
\multicolumn{1}{c}{}                         & SLIP~\citep{MuK0X22}          & 41.17       & {81.30}  & {53.68}                   \\
\multicolumn{1}{c}{}                   \cellcolor{LightCyan}     & SLIP-\hclip  & \cellcolor{LightCyan}\textbf{45.64}      & \cellcolor{LightCyan}\textbf{{82.31}}  & \cellcolor{LightCyan}\textbf{{53.79}}                   \\
\midrule
\multicolumn{1}{c}{\multirow{4}{*}{\rotatebox{90}{YFCC15M}}} 
& SLIP~\citep{MuK0X22}          & 25.29 (34.30$^{*}$)    & 60.19 &   26.80                \\
\multicolumn{1}{c}{} \cellcolor{LightCyan} & SLIP-\hclip   & \cellcolor{LightCyan}35.43      &  \cellcolor{LightCyan}75.49       &   \cellcolor{LightCyan}47.84                        \\
\multicolumn{1}{c}{}                         & DECLIP~\citep{LiLZCOSYY22}        & 36.05 (43.20$^{*}$) &  78.12       &     50.60                      \\
\multicolumn{1}{c}{}    \cellcolor{LightCyan}  & DECLIP-\hclip & \cellcolor{LightCyan}\textbf{43.80}    &  \cellcolor{LightCyan} \textbf{84.88}      &   \cellcolor{LightCyan}  \textbf{56.31}          \\ 
\midrule
\multicolumn{1}{c}{\multirow{3}{*}{\rotatebox{90}{\tmlrnew{29M}}}} & \tmlrnew{CLIP$^{\dagger}$~\citep{RadfordKHRGASAM21}}          & \tmlrnew{42.32}    & \tmlrnew{71.98} &   \tmlrnew{42.73}                \\
\multicolumn{1}{c}{} \cellcolor{Gray} & \tmlrnew{CLIP$^{\dagger}$ Cont. Train}   & \cellcolor{Gray} \tmlrnew{{42.25}}     &  \cellcolor{Gray}\tmlrnew{{71.72}}       &   \cellcolor{Gray}\tmlrnew{{42.66}}                        \\
\multicolumn{1}{c}{} \cellcolor{LightCyan} & \tmlrnew{CLIP$^{\dagger}$-\hclip}   & \cellcolor{LightCyan} \tmlrnew{\textbf{46.33}}     &  \cellcolor{LightCyan}\tmlrnew{\textbf{77.97}}       &   \cellcolor{LightCyan}\tmlrnew{\textbf{48.33}}                        \\
\bottomrule
\end{tabular}
}
}
\caption{
Zero-shot classification performance on ImageNet, CIFAR-10, and CIFAR-100. Baselines marked with ${\dagger}$ were trained by us; others use publicly available pre-trained parameters. For SLIP and DECLIP on YFCC15M, we report both our evaluation using OpenCLIP with pre-trained parameters from \cite{LiLZCOSYY22} and the results reported in \cite{LiLZCOSYY22}, marked with ${*}$.
\label{tab:main_table}}
\vspace{-20pt}
\end{table}

\noindent\textbf{Zero-Shot Fine-Grained Classification.} Utilizing hard image-text pairs in contrastive learning, \hclip\ enhances the discriminative power of CLIP's visual embeddings, benefiting fine-grained classification tasks. As shown in Table \ref{tab:fine_grained_zs}, SLIP-\hclip\ improves zero-shot accuracy on Caltech101 by 12.88\% and 3.95\% for models pre-trained on CC3M and CC12M, respectively. Both CLIP and SLIP models consistently improve when augmented with \hclip. The above results indicate that the embedding space becomes tighter when using hard pairs in contrastive loss.
\begin{table*}[h]
    \centering
        \small
    \setlength{\tabcolsep}{3.5pt}{
    \begin{tabular}{cccccccccc}
        \toprule
    \small
    \shortstack[c]{Dataset} &
    \shortstack[c]{Method} &
    \rotatebox{0}{\footnotesize{Caltech101}} &
    \rotatebox{0}{\footnotesize{Food101}} &
    \rotatebox{0}{\footnotesize{Sun397}} &
    \rotatebox{0}{\footnotesize{Flowers102}} & 
    \rotatebox{0}{\footnotesize{CUB}} &
    \rotatebox{0}{\footnotesize{Stanford Cars}} &
    \rotatebox{0}{\footnotesize{FGVC Aircraft}} &
    \rotatebox{0}{\footnotesize{Average}}\\
    \toprule
    \multirow{4}{*}{CC3M}     & CLIP & 42.14 & 13.02 & 27.08 & 13.37 & 3.45 & 1.08 & 1.02 & 14.45    \\
    & CLIP-\hclip \cellcolor{LightCyan} & \cellcolor{LightCyan}48.08 & \cellcolor{LightCyan}13.11 & \cellcolor{LightCyan}28.94 & \cellcolor{LightCyan}13.61 & \cellcolor{LightCyan}3.70 & \cellcolor{LightCyan}1.17 & \cellcolor{LightCyan}1.11 & \cellcolor{LightCyan}15.67  \\
    & SLIP & 54.01 & 16.03 & 29.19 & 12.06 & 4.70 & \textbf{1.21} & \textbf{1.50} & 16.96  \\
    & SLIP-\hclip \cellcolor{LightCyan} & \cellcolor{LightCyan}\textbf{66.89} & \cellcolor{LightCyan}\textbf{17.05} & \cellcolor{LightCyan}\textbf{33.69} & \cellcolor{LightCyan}\textbf{15.16} & \cellcolor{LightCyan}\textbf{4.85} & \cellcolor{LightCyan}1.19 & \cellcolor{LightCyan} 1.29 & \cellcolor{LightCyan} \textbf{20.12}  \\
    \midrule
    \multirow{4}{*}{CC12M}& CLIP & 63.78 & 31.53 & 37.86 & 19.56 & 7.32 & 14.22 & 2.49 & 25.25  \\
    & CLIP-\hclip \cellcolor{LightCyan} & \cellcolor{LightCyan}64.85 & \cellcolor{LightCyan}36.49 & \cellcolor{LightCyan}38.22 & \cellcolor{LightCyan}24.73 & \cellcolor{LightCyan}8.58 & \cellcolor{LightCyan}15.59 & \cellcolor{LightCyan}2.97 & \cellcolor{LightCyan}27.35  \\
    & SLIP & 76.33 & 52.33 & 44.96 & \textbf{31.81} & 10.50 & 22.53 & 3.06 & 34.50   \\
    & SLIP-\hclip \cellcolor{LightCyan} & \cellcolor{LightCyan}\textbf{80.28} & \cellcolor{LightCyan}\textbf{54.86} & \cellcolor{LightCyan}\textbf{47.53} & \cellcolor{LightCyan}31.39 & \cellcolor{LightCyan}\textbf{10.56} & \cellcolor{LightCyan}\textbf{25.67} & \cellcolor{LightCyan}\textbf{4.08} & \cellcolor{LightCyan}\textbf{36.34}  \\
    \bottomrule
    \end{tabular}
    }
    \caption{{Zero-shot performance on fine-grained image classification.}  On a variety of fine-grained classification benchmarks, \hclip\ consistent boosts the model performance compared to the original versions.}
    \label{tab:fine_grained_zs}
\vspace{-10pt}
\end{table*}

\noindent\textbf{Zero-Shot Retrieval.} We evaluate \hclip\  on zero-shot image-to-text retrieval tasks on MS-COCO ~\citep{DBLP:conf/eccv/LinMBHPRDZ14} and Flickr30K ~\citep{DBLP:conf/iccv/PlummerWCCHL15}. As shown Table~\ref{tab:retrieval}, both CLIP and SLIP, pre-trained on CC3M and CC12M , consistently improved by \hclip.
\begin{table}[!h]
    \centering
    \small
    \setlength{\tabcolsep}{3pt}{
    \begin{tabular}{cccccc}
    \toprule
     \multirow{2}{*}{\shortstack[c]{Pretraining\\ Dataset}} & 
     \multirow{2}{*}{Method} & 
     \multicolumn{2}{c}{COCO} &
     \multicolumn{2}{c}{Flickr30K} \\
    && \footnotesize{R@1} $\uparrow$ & \footnotesize{R@5} $\uparrow$ & \footnotesize{R@1} $\uparrow$ & \footnotesize{R@5} $\uparrow$ 
       \\
\toprule

    \multirow{4}{*}{CC3M}  & CLIP & 14.4 & 34.1 & 31.7 & 56.0 \\
                           & CLIP-\hclip \cellcolor{LightCyan} & \cellcolor{LightCyan}17.8 & \cellcolor{LightCyan}39.8 & \cellcolor{LightCyan}35.4 & \cellcolor{LightCyan}61.0   \\
                            & SLIP & 22.3 & 45.6 & 39.6 & 68.6 \\
                           & SLIP-\hclip \cellcolor{LightCyan} & \cellcolor{LightCyan}\textbf{23.4} & \cellcolor{LightCyan}\textbf{48.3} & \cellcolor{LightCyan}\textbf{41.8} & \cellcolor{LightCyan}\textbf{69.6} \\

    \midrule
    
    \multirow{4}{*}{CC12M} & CLIP & 26.9 & 52.6 & 47.2 & 74.3  \\
                           & CLIP-\hclip \cellcolor{LightCyan} & \cellcolor{LightCyan}27.8 & \cellcolor{LightCyan}54.3 & \cellcolor{LightCyan}48.2 & \cellcolor{LightCyan}75.4 \\
                          & SLIP & 39.0 & 66.0 & 65.4 & \textbf{90.1}\\
                           & SLIP-\hclip \cellcolor{LightCyan} & \cellcolor{LightCyan}\textbf{39.4} & \cellcolor{LightCyan}\textbf{67.2} & \cellcolor{LightCyan}\textbf{66.2} & \cellcolor{LightCyan}89.7 \\
    
    \bottomrule
    \end{tabular}
    }
     \vspace{-2pt}
    \caption{{Zero-shot image-text retrieval results on MSCOCO and Flickr.  } $\uparrow$ indicates higher is better. 
    \label{tab:retrieval}
	}
 \vspace{-25pt}
\end{table}

\noindent\textbf{Linear Probing.}
The linear probing task trains a randomly initialized linear classifier on the feature extracted from the frozen image encoder on the downstream dataset. We train the logistic regression classifier using scikit-learn’s L-BFGS implementation~\citep{pedregosa2011scikit}, with maximum 1,000 iterations on those 7 datasets. For each dataset, we search for the best regularization strength factor on the validation set over 45 logarithmically spaced steps within the range 1e-6 to 1e+5.
Experimental results in Table~\ref{tab:linear_probe} demonstrate that both CLIP-\hclip\ and SLIP-\hclip\ have consistent improvements over their counterparts on almost all 7 datasets. Note that on CC12M, SLIP-\hclip\ performs marginally better on 5 out of 7 datasets. It's probably because the self-supervision of SLIP~\citep{MuK0X22} within the visual modal can be beneficial for learning fine-grained visual embedding, while SLIP-\hclip\ doesn't include image self-supervision during the training. And we did not match the training batch size as SLIP~\citep{MuK0X22} because of resource limitations. A combination of \hclip\ and image self-supervision with larger batch size may be a potential direction for achieving better linear probe performance.
\begin{table*}[h]
    \centering
    \small
    \setlength{\tabcolsep}{4pt}{
    \begin{tabular}{cccccccccc}
        \toprule
    \shortstack[c]{Dataset} &
    \shortstack[c]{Method} &
    \rotatebox{0}{\footnotesize{Caltech101}} &
    \rotatebox{0}{\footnotesize{Food101}} &
    \rotatebox{0}{\footnotesize{Sun397}} &
    \rotatebox{0}{\footnotesize{Flowers102}} & 
    \rotatebox{0}{\footnotesize{CUB}} &
    \rotatebox{0}{\footnotesize{Stanford Cars}} &
    \rotatebox{0}{\footnotesize{FGVC Aircraft}} &
    \rotatebox{0}{\footnotesize{Avg.}}\\
    \toprule
    \multirow{5}{*}{{ CC3M}}     & { CYCLIP} & 80.88 & 54.95 & - & 83.74 & - & 22.72 & 28.02 & -   \\
    & { CLIP} &  80.11 & 53.82 & 56.40 & 84.07 & 40.30 & 22.70 & 35.61 & 53.29\\
    & CLIP-\hclip \cellcolor{LightCyan} & \cellcolor{LightCyan} 82.49 & \cellcolor{LightCyan} 59.79 & \cellcolor{LightCyan} 59.56 & \cellcolor{LightCyan} 87.84 & \cellcolor{LightCyan} 46.19 & \cellcolor{LightCyan} 30.01 & \cellcolor{LightCyan} 42.48 & \cellcolor{LightCyan} 58.34  \\
    & { SLIP} & 87.96 &  72.50  & 66.96  & 91.91 & 49.77 & 39.25 & 45.87 & 64.89 \\
    & SLIP-\hclip \cellcolor{LightCyan} & \cellcolor{LightCyan}\textbf{89.64} & \cellcolor{LightCyan}\textbf{73.09} & \cellcolor{LightCyan}\textbf{67.67} & \cellcolor{LightCyan}\textbf{93.02} & \cellcolor{LightCyan}\textbf{53.16} & \cellcolor{LightCyan}\textbf{42.44}& \cellcolor{LightCyan}\textbf{48.66} &  \cellcolor{LightCyan}\textbf{66.81} \\
    \midrule
    \multirow{4}{*}{{ CC12M}}     & { CLIP} & 85.35 & 68.00 & 64.45 & 87.88 & 48.75 & 57.80 & 40.32 & 64.65 \\
    & CLIP-\hclip \cellcolor{LightCyan} & \cellcolor{LightCyan}85.87 & \cellcolor{LightCyan}68.89 & \cellcolor{LightCyan}64.95 & \cellcolor{LightCyan}88.36 & \cellcolor{LightCyan}49.41 & \cellcolor{LightCyan}58.55 & \cellcolor{LightCyan}40.17 & \cellcolor{LightCyan}65.17  \\
    & SLIP & \textbf{92.89}  &  83.63  & 74.34   & 94.87 & \textbf{60.99} & 73.43 & 52.23 & 76.05 \\
    & SLIP-\hclip \cellcolor{LightCyan} & \cellcolor{LightCyan}92.85 & \cellcolor{LightCyan}\textbf{84.25} & \cellcolor{LightCyan}\textbf{74.74} & \cellcolor{LightCyan}\textbf{95.09} & \cellcolor{LightCyan}60.53 & \cellcolor{LightCyan}\textbf{74.23} & \cellcolor{LightCyan}\textbf{52.36} & \cellcolor{LightCyan}\textbf{76.29}  \\
    \bottomrule
    \end{tabular}
    }
     \vspace{-3pt}
    \caption{{Linear probe performance on Fine-grained Image Classification.} On average, the linear probe performance of CLIP and SLIP pretrained on CC3M and CC12M are improved.}
    \label{tab:linear_probe}
\vspace{-1pt}
\end{table*}

\subsection{HELIP with Scaled Training Data}
\label{subsec:scaled_exp}
\begin{figure}[h]
\vspace{-5pt}
\begin{center}
\small
\includegraphics[width=\linewidth]{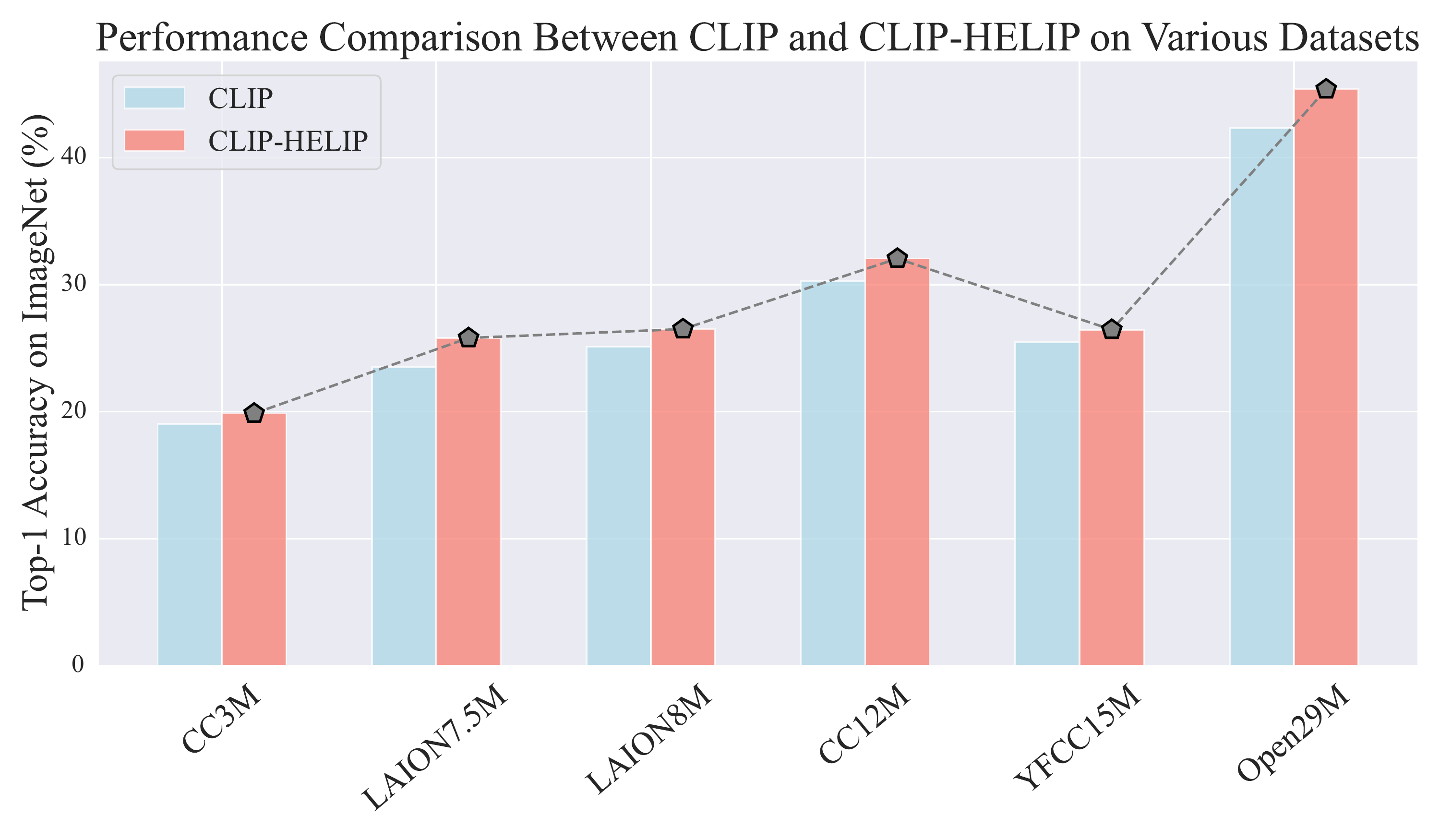}
\vspace{-10pt}
\caption{
\label{fig:scaleup}
{Zero-shot performance on ImageNet for models pre-trained on different dataset sizes}.
}
\end{center}
\vspace{-5pt}
\end{figure}

\vspace{-2pt}
To investigate the impact of expanded training dataset sizes on the effectiveness of \hclip, we trained the CLIP model on the YFCC15M dataset. This training yielded a zero-shot classification accuracy of 25.46\% on ImageNet. After applying \hclip and one epoch of training, its performance improved to 26.45\%. To summarize the zero-shot performance on ImageNet of both the standard CLIP and its enhanced version, CLIP-\hclip, across different data scales, we have illustrated these results in Figure~\ref{fig:scaleup}.
The results show that \hclip\ consistently enhances CLIP's performance. Most notably, the largest dataset, Open29M, witnessed a remarkable performance increase of 3.06\% with \hclip. This result indicates that \hclip\ can provide immediate performance enhancements for well-trained CLIP models on larger datasets, such as the private 400M dataset mentioned in~\cite{RadfordKHRGASAM21}.

\begin{figure}[h]
\centering
\includegraphics[width=\linewidth]{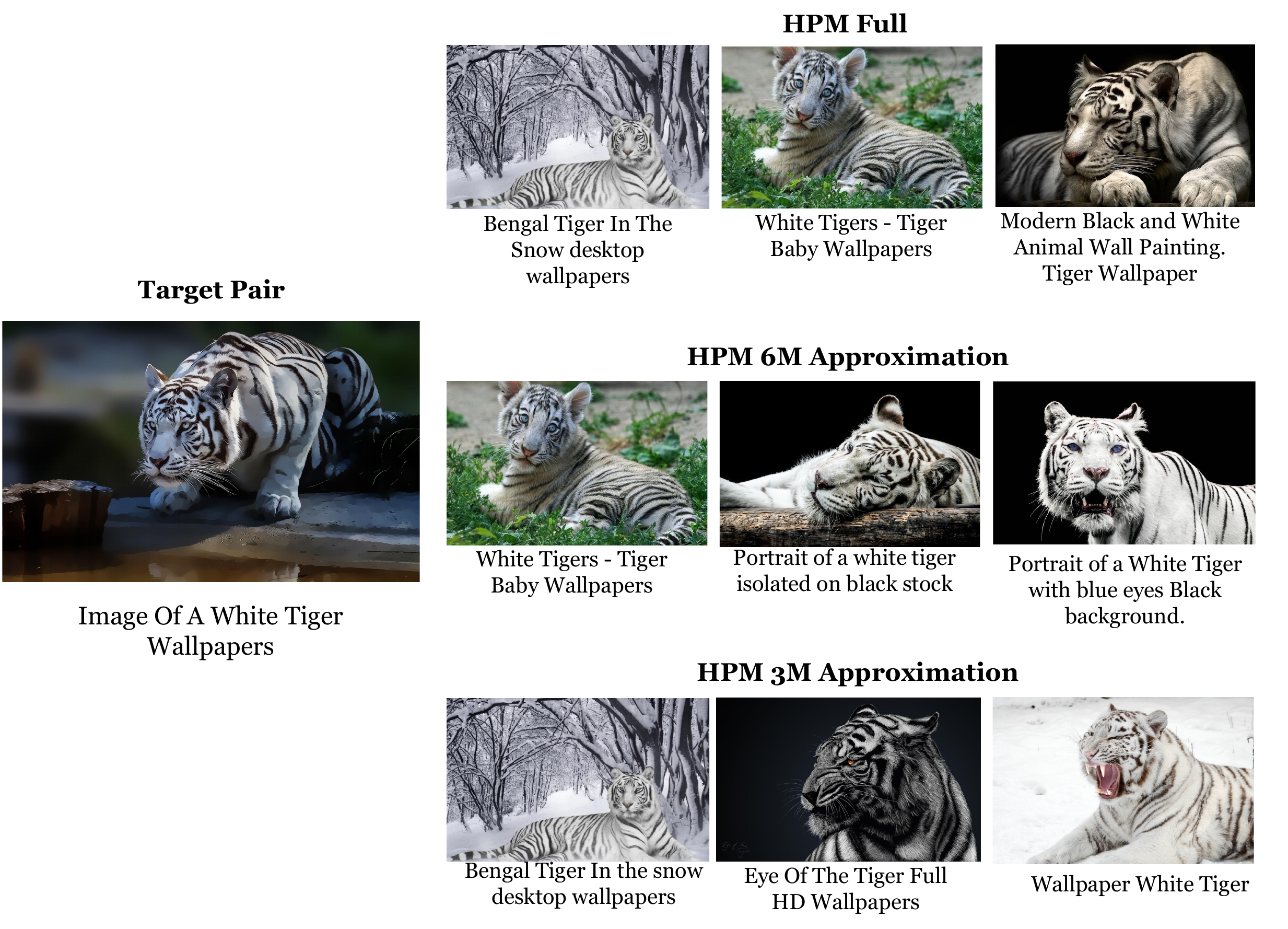}
\vspace{-5pt}
\caption{{Hard pairs} from HPM and fastHPM. FastHPM produces high-quality {hard pairs} that compete with HPM.}
\vspace{-5pt}
\label{fig:approximation_vis}
\end{figure}

\subsection{Impact of Hard Negative Margin Loss}
\label{subsec:margin_loss}
We examine the impact of hard negative margin loss (HNML) on the SLIP model's performance. Specifically, we analyze the SLIP model pre-trained on CC3M when further trained with both HPM and HNML, compared to training without HNML. We evaluated zero-shot classification performance across datasets including ImageNet, CIFAR-100, CIFAR-10, Caltech-101, Food-101, and SUN397.
As detailed in Table~\ref{tab:aba_margin_loss}, the SLIP model augmented with HPM and HNML outperforms the baseline SLIP and SLIP + HPM models by 4.51 and 3.27 points, respectively. Notably, the model achieves better performance on CIFAR-10 without HNML. We hypothesize that HNML enhances the discriminative power of learned representations by incorporating class distance as a cost metric. Therefore, for classification tasks with a larger number of subclasses, employing HNML during training can improve performance.
\begin{table}[h]
\centering
\scalebox{0.7}{
\setlength{\tabcolsep}{1pt}{
\begin{tabular}{cccccccc}
\toprule
     & ImageNet & CF10 & CF100 &  Caltech101 & Food101 & Sun397  & Avg. \\ 
\toprule
SLIP   &  23.00  &  65.61  & 34.69  &  54.01 &  16.03 & 29.20  & 37.09 \\
wo HNML &  24.94    &   \textbf{69.44}   & 36.35   &  64.07 &  16.51 & 30.91 &   40.37  \\
w HNML &  \textbf{26.05}  &  68.18   &  \textbf{37.77}  &  \textbf{66.89} &  \textbf{17.05} & \textbf{33.68} & \textbf{41.60} \\
\bottomrule
\end{tabular}
}
}
\caption{
{SLIP continous training with and without hard negative margin loss.}  
\label{tab:aba_margin_loss}
} 
\vspace{-15pt}
\end{table}


\subsection{Delving into Hard Pair Mining}
\label{subsec:hpm_detail}
\noindent\textbf{Performance Comparison between HPM and FastHPM.}
A comparison was made between the zero-shot performances of SLIP models, further trained with \tmlrnew{hard pairs} obtained from both HPM and fastHPM. This comparison, conducted under three different settings, was summarized in Table~\ref{tab:approx}. Additionally, we established subsets $\widetilde{\mathcal{D}}_i$ of sizes 3M and 6M, and accordingly denoted \hclip\ with these subset sizes as \hclip-3M and \hclip-6M.
Table~\ref{tab:approx} shows that the zero-shot performances of \hclip-3M and \hclip-6M remain competitive with the global HPM  \tmlrnew{hard pair mining} approach. 
These findings suggest that fastHPM offers an efficient hard pair mining strategy without compromising performance and has the potential to scale up in larger pre-training datasets.
\begin{table}[h]
\vspace{-5pt}
\centering
\small
\setlength{\tabcolsep}{8pt}{
\begin{tabular}{cccc}
\toprule
 & Imagenet & CIFAR10 & CIFAR100 \\ 
\toprule
SLIP & 41.17 & 81.30 & 53.68 \\
\hclip - 3M    &    45.07               &   \textbf{82.42}               &      55.22             \\
\hclip - 6M   &   44.98                &     81.64             &       \textbf{56.62}            \\
\hclip - Full  &   \textbf{45.64}         &      82.31       &   53.79         \\ 
\bottomrule
\end{tabular}
}
\vspace{-5pt}
\caption{{Zero-shot performance of SLIP with HELIP on CC12M with hard pairs from HPM and fastHPM.} 
\label{tab:approx}
}
\vspace{-5pt}
\end{table}

\noindent\textbf{Visual insights into HPM and FastHPM.}
We took the initiative to visualize the \tmlrnew{hard pairs} as identified by the aforementioned three methods. Within Figure~\ref{fig:approximation_vis}, the leftmost image-text pairing is earmarked as the target. The pairs in the primary row are those selected via HPM. The subsequent rows, specifically the second and third, present image-text pairings identified by the 6M fastHPM and the 3M fastHPM methods, respectively. Through a comparative visualization, it's evident that the \tmlrnew{hard pairs} pinpointed by fastHPM bear a significant resemblance to the target pair. For readers keen on delving deeper, we've provided an extended set of visualization outcomes in Appendix~\ref{apd:more_exp}.

\noindent\textbf{Computational Time Analysis.}
Table~\ref{tab:time_stat} provides a comparison of the computational time required by HPM and fastHPM. The \tmlrnew{hard negative pairs} preparation times listed were measured on 8 V100 GPUs, with the exception of the $*$ symbol, which was measured on a single V100 GPU.
Given its efficiency and the performance similarities observed in Table~\ref{tab:approx}, fastHPM emerges as a compelling alternative to the full HPM method.
\begin{table}[h]
    \centering
    \small
    \setlength{\tabcolsep}{7pt}{
    \begin{tabular}{cccc}
    \toprule
                 & CC3M & CC12M & YFCC15M \\ 
    \toprule
    \hclip - 3M    &    -       &   2h18min      &   3h27min      \\
    \hclip - 6M   &     -     &      5h3min    &    6h19min    \\
    \hclip - Full  &     1h9min$^{*}$       &     9h11min     &   17h41min     \\ 
    \bottomrule
    \end{tabular}
    }
    \caption{{Preparation time for hard pairs.} FastHPM speeds up the \tmlrnew{hard negative pairs} mining process.\label{tab:time_stat}}
\end{table}
\section{Conclusion}
\label{sec:conclusion}
In this work, we present \hclip, a framework that enhances pre-trained CLIP models by more effectively utilizing their original training datasets. \hclip offers a cost-effective and easily integrable solution for improving existing models without extensive retraining or additional data. Specifically, it treats each text-image pair as a point in the joint vision-language space and identifies hard pairs, those that are close together, using the Hard Pair Mining (HPM) strategy. Furthermore, to efficiently leverage these hard pairs, we introduce the Hard Negative Margin Loss (HNML). Empirically, we found that \hclip boosts the performance of existing checkpoints within a few epochs of continuous training. Evaluations across various benchmarks, including zero-shot classification, image-text retrieval, and linear probing, demonstrate the effectiveness and efficiency of our method.
These findings highlight that in the era of large-scale models and datasets, performance improvement can be achieved not only by collecting more data or scaling up models, but also by intelligently maximizing the utility of the data we already have.

\section{Limitations}
\label{sec:limitation}
While \hclip\ demonstrates significant improvements over existing CLIP models, several limitations should be acknowledged.

\paragraph{Dependence on Dataset Quality and Diversity.} The effectiveness of \hclip\ is inherently tied to the quality and diversity of the original training dataset. If the dataset contains limited variability or is biased toward certain domains, the benefits of mining hard pairs may diminish. In scenarios where datasets are very small, the selection of meaningful challenging pairs becomes more difficult, potentially limiting the overall performance gains.

\paragraph{Reusing Original Training Data.} \hclip\ leverages the original pretraining data without incorporating additional external data sources. While this approach reduces the need for costly data collection, it might also inherits biases present in the original dataset.

\paragraph{Limited Evaluation on Diverse Datasets.} The empirical evaluations of \hclip\ are conducted on specific datasets like CC3M, CC12M, YFCC15M and subsets of LAION. While these are widely accepted benchmarks, the generalizability of its improvements to real-world applications requires further evaluation.

\section{Acknowledgments}
This material is based upon work supported by the Air Force Office of Scientific Research under 
award number FA2386-24-1-4011. This research is also partially supported by the Singapore Ministry of Education Academic Research Fund Tier 1 (Award No.\ T1 251RES2207). 

Additional support for Minbin Huang and Hong Cheng is provided by project \#MMT-p2-23 of the Shun Hing Institute of Advanced Engineering at The Chinese University of Hong Kong, and by the Research Grants Council of the Hong Kong Special Administrative Region, China (No.\ CUHK 14217622).

The authors gratefully acknowledge all of these sources of support. Any opinions, findings, and 
conclusions or recommendations expressed in this publication are those of the authors and do not 
necessarily reflect the views of the funding agencies.

\bibliography{custom}

\begin{thebibliography}{51}
\providecommand{\natexlab}[1]{#1}

\bibitem[{Awais et~al.(2023)Awais, Naseer, Khan, Anwer, Cholakkal, Shah, Yang, and Khan}]{awais2023foundational}
Muhammad Awais, Muzammal Naseer, Salman Khan, Rao~Muhammad Anwer, Hisham Cholakkal, Mubarak Shah, Ming-Hsuan Yang, and Fahad~Shahbaz Khan. 2023.
\newblock Foundational models defining a new era in vision: A survey and outlook.
\newblock \emph{arXiv preprint arXiv:2307.13721}.

\bibitem[{Baldrati et~al.(2022)Baldrati, Bertini, Uricchio, and Bimbo}]{Baldrati_2022_CVPR}
Alberto Baldrati, Marco Bertini, Tiberio Uricchio, and Alberto~Del Bimbo. 2022.
\newblock \href {https://doi.org/10.1109/CVPRW56347.2022.00543} {Conditioned and composed image retrieval combining and partially fine-tuning clip-based features}.
\newblock In \emph{{IEEE/CVF} Conference on Computer Vision and Pattern Recognition Workshops, {CVPR} Workshops 2022, New Orleans, LA, USA, June 19-20, 2022}, pages 4955--4964. {IEEE}.

\bibitem[{Bossard et~al.(2014)Bossard, Guillaumin, and Van~Gool}]{bossard2014food}
Lukas Bossard, Matthieu Guillaumin, and Luc Van~Gool. 2014.
\newblock Food-101--mining discriminative components with random forests.
\newblock In \emph{European conference on computer vision}, pages 446--461. Springer.

\bibitem[{Brown et~al.(2021)Brown, Bun, Feldman, Smith, and Talwar}]{brown2021memorization}
Gavin Brown, Mark Bun, Vitaly Feldman, Adam Smith, and Kunal Talwar. 2021.
\newblock When is memorization of irrelevant training data necessary for high-accuracy learning?
\newblock In \emph{Proceedings of the 53rd annual ACM SIGACT symposium on theory of computing}, pages 123--132.

\bibitem[{Cai et~al.(2020)Cai, Frankle, Schwab, and Morcos}]{negativesnotequal}
Tiffany~Tianhui Cai, Jonathan Frankle, David~J. Schwab, and Ari~S. Morcos. 2020.
\newblock Are all negatives created equal in contrastive instance discrimination?
\newblock \emph{ArXiv preprint}.

\bibitem[{Caron et~al.(2021)Caron, Touvron, Misra, J{\'{e}}gou, Mairal, Bojanowski, and Joulin}]{caron2021emerging}
Mathilde Caron, Hugo Touvron, Ishan Misra, Herv{\'{e}} J{\'{e}}gou, Julien Mairal, Piotr Bojanowski, and Armand Joulin. 2021.
\newblock \href {https://doi.org/10.1109/ICCV48922.2021.00951} {Emerging properties in self-supervised vision transformers}.
\newblock In \emph{2021 {IEEE/CVF} International Conference on Computer Vision, {ICCV} 2021, Montreal, QC, Canada, October 10-17, 2021}, pages 9630--9640. {IEEE}.

\bibitem[{Changpinyo et~al.(2021)Changpinyo, Sharma, Ding, and Soricut}]{changpinyo2021conceptual}
Soravit Changpinyo, Piyush Sharma, Nan Ding, and Radu Soricut. 2021.
\newblock \href {https://doi.org/10.1109/CVPR46437.2021.00356} {Conceptual 12m: Pushing web-scale image-text pre-training to recognize long-tail visual concepts}.
\newblock In \emph{{IEEE} Conference on Computer Vision and Pattern Recognition, {CVPR} 2021, virtual, June 19-25, 2021}, pages 3558--3568. Computer Vision Foundation / {IEEE}.

\bibitem[{Chen et~al.(2022)Chen, Chen, Xu, and Xu}]{chen2022improving}
Feilong Chen, Xiuyi Chen, Shuang Xu, and Bo~Xu. 2022.
\newblock Improving cross-modal understanding in visual dialog via contrastive learning.
\newblock In \emph{ICASSP 2022-2022 IEEE International Conference on Acoustics, Speech and Signal Processing (ICASSP)}, pages 7937--7941. IEEE.

\bibitem[{Chen et~al.(2020{\natexlab{a}})Chen, Kornblith, Norouzi, and Hinton}]{chen2020simple}
Ting Chen, Simon Kornblith, Mohammad Norouzi, and Geoffrey~E. Hinton. 2020{\natexlab{a}}.
\newblock \href {http://proceedings.mlr.press/v119/chen20j.html} {A simple framework for contrastive learning of visual representations}.
\newblock In \emph{Proc. of ICML}, volume 119 of \emph{Proceedings of Machine Learning Research}, pages 1597--1607. {PMLR}.

\bibitem[{Chen et~al.(2020{\natexlab{b}})Chen, Fan, Girshick, and He}]{mocov2}
Xinlei Chen, Haoqi Fan, Ross~B. Girshick, and Kaiming He. 2020{\natexlab{b}}.
\newblock Improved baselines with momentum contrastive learning.
\newblock \emph{ArXiv preprint}.

\bibitem[{Chen et~al.(2009)Chen, Garcia, Gupta, Rahimi, and Cazzanti}]{chen2009similarity}
Yihua Chen, Eric~K Garcia, Maya~R Gupta, Ali Rahimi, and Luca Cazzanti. 2009.
\newblock Similarity-based classification: Concepts and algorithms.
\newblock \emph{Journal of Machine Learning Research}, 10(3).

\bibitem[{Cui et~al.(2022)Cui, Zhao, Liang, Li, and Shao}]{cui2022democratizing}
Yufeng Cui, Lichen Zhao, Feng Liang, Yangguang Li, and Jing Shao. 2022.
\newblock \href {https://arxiv.org/abs/2203.05796} {Democratizing contrastive language-image pre-training: A clip benchmark of data, model, and supervision}.
\newblock \emph{Preprint}, arXiv:2203.05796.

\bibitem[{Deng et~al.(2009)Deng, Dong, Socher, Li, Li, and Li}]{DBLP:conf/cvpr/DengDSLL009}
Jia Deng, Wei Dong, Richard Socher, Li{-}Jia Li, Kai Li, and Fei{-}Fei Li. 2009.
\newblock \href {https://doi.org/10.1109/CVPR.2009.5206848} {Imagenet: {A} large-scale hierarchical image database}.
\newblock In \emph{2009 {IEEE} Computer Society Conference on Computer Vision and Pattern Recognition {(CVPR} 2009), 20-25 June 2009, Miami, Florida, {USA}}, pages 248--255. {IEEE} Computer Society.

\bibitem[{Fei-Fei et~al.(2004)Fei-Fei, Rob, and Perona}]{fei2004learning}
Li~Fei-Fei, Fergus Rob, and Pietro Perona. 2004.
\newblock Learning generative visual models from few training examples: An incremental bayesian approach tested on 101 object categories.
\newblock In \emph{Computer Vision and Pattern Recognition Workshop, 2004. CVPRW'04. Conference on}. IEEE.

\bibitem[{F{\"{u}}rst et~al.(2021)F{\"{u}}rst, Rumetshofer, Tran, Ramsauer, Tang, Lehner, Kreil, Kopp, Klambauer, Bitto{-}Nemling, and Hochreiter}]{CLOOB}
Andreas F{\"{u}}rst, Elisabeth Rumetshofer, Viet Tran, Hubert Ramsauer, Fei Tang, Johannes Lehner, David~P. Kreil, Michael Kopp, G{\"{u}}nter Klambauer, Angela Bitto{-}Nemling, and Sepp Hochreiter. 2021.
\newblock {CLOOB:} modern hopfield networks with infoloob outperform {CLIP}.
\newblock \emph{ArXiv preprint}.

\bibitem[{Gadre et~al.(2023)Gadre, Ilharco, Fang, Hayase, Smyrnis, Nguyen, Marten, Wortsman, Ghosh, Zhang et~al.}]{gadre2023datacomp}
Samir~Yitzhak Gadre, Gabriel Ilharco, Alex Fang, Jonathan Hayase, Georgios Smyrnis, Thao Nguyen, Ryan Marten, Mitchell Wortsman, Dhruba Ghosh, Jieyu Zhang, et~al. 2023.
\newblock Datacomp: In search of the next generation of multimodal datasets.
\newblock \emph{ArXiv preprint}.

\bibitem[{Goel et~al.(2022)Goel, Bansal, Bhatia, Rossi, Vinay, and Grover}]{goel2022cyclip}
Shashank Goel, Hritik Bansal, Sumit Bhatia, Ryan~A Rossi, Vishwa Vinay, and Aditya Grover. 2022.
\newblock Cyclip: Cyclic contrastive language-image pretraining.
\newblock \emph{ArXiv preprint}.

\bibitem[{He et~al.(2022)He, Zhou, Ma, Berg{-}Kirkpatrick, and Neubig}]{he2022towards}
Junxian He, Chunting Zhou, Xuezhe Ma, Taylor Berg{-}Kirkpatrick, and Graham Neubig. 2022.
\newblock \href {https://openreview.net/forum?id=0RDcd5Axok} {Towards a unified view of parameter-efficient transfer learning}.
\newblock In \emph{Proc. of ICLR}. OpenReview.net.

\bibitem[{Huynh et~al.(2022)Huynh, Kornblith, Walter, Maire, and Khademi}]{HuynhKWMK22}
Tri Huynh, Simon Kornblith, Matthew~R. Walter, Michael Maire, and Maryam Khademi. 2022.
\newblock Boosting contrastive self-supervised learning with false negative cancellation.
\newblock In \emph{IEEE/CVF Winter Conference on Applications of Computer Vision, WACV 2022, Waikoloa, HI, USA, January 3-8, 2022}.

\bibitem[{Jia et~al.(2021)Jia, Yang, Xia, Chen, Parekh, Pham, Le, Sung, Li, and Duerig}]{JiaYXCPPLSLD21}
Chao Jia, Yinfei Yang, Ye~Xia, Yi{-}Ting Chen, Zarana Parekh, Hieu Pham, Quoc~V. Le, Yun{-}Hsuan Sung, Zhen Li, and Tom Duerig. 2021.
\newblock \href {http://proceedings.mlr.press/v139/jia21b.html} {Scaling up visual and vision-language representation learning with noisy text supervision}.
\newblock In \emph{Proc. of ICML}, volume 139 of \emph{Proceedings of Machine Learning Research}, pages 4904--4916. {PMLR}.

\bibitem[{Kalantidis et~al.(2020)Kalantidis, Sariyildiz, Pion, Weinzaepfel, and Larlus}]{KalantidisSPWL20}
Yannis Kalantidis, Mert~B{\"{u}}lent Sariyildiz, No{\'{e}} Pion, Philippe Weinzaepfel, and Diane Larlus. 2020.
\newblock \href {https://proceedings.neurips.cc/paper/2020/hash/f7cade80b7cc92b991cf4d2806d6bd78-Abstract.html} {Hard negative mixing for contrastive learning}.
\newblock In \emph{Advances in Neural Information Processing Systems 33: Annual Conference on Neural Information Processing Systems 2020, NeurIPS 2020, December 6-12, 2020, virtual}.

\bibitem[{Krause et~al.(2013)Krause, Stark, Deng, and Fei-Fei}]{krause20133d}
Jonathan Krause, Michael Stark, Jia Deng, and Li~Fei-Fei. 2013.
\newblock 3d object representations for fine-grained categorization.
\newblock In \emph{Proceedings of the IEEE International Conference on Computer Vision Workshops}, pages 554--561.

\bibitem[{Krizhevsky et~al.(2009)Krizhevsky, Hinton et~al.}]{krizhevsky2009learning}
Alex Krizhevsky, Geoffrey Hinton, et~al. 2009.
\newblock Learning multiple layers of features from tiny images.

\bibitem[{Li et~al.(2022{\natexlab{a}})Li, Li, Xiong, and Hoi}]{0001LXH22}
Junnan Li, Dongxu Li, Caiming Xiong, and Steven C.~H. Hoi. 2022{\natexlab{a}}.
\newblock \href {https://proceedings.mlr.press/v162/li22n.html} {{BLIP:} bootstrapping language-image pre-training for unified vision-language understanding and generation}.
\newblock In \emph{International Conference on Machine Learning, {ICML} 2022, 17-23 July 2022, Baltimore, Maryland, {USA}}, volume 162 of \emph{Proceedings of Machine Learning Research}, pages 12888--12900. {PMLR}.

\bibitem[{Li et~al.(2021)Li, Selvaraju, Gotmare, Joty, Xiong, and Hoi}]{LiSGJXH21}
Junnan Li, Ramprasaath~R. Selvaraju, Akhilesh Gotmare, Shafiq~R. Joty, Caiming Xiong, and Steven~Chu{-}Hong Hoi. 2021.
\newblock \href {https://proceedings.neurips.cc/paper/2021/hash/505259756244493872b7709a8a01b536-Abstract.html} {Align before fuse: Vision and language representation learning with momentum distillation}.
\newblock In \emph{Advances in Neural Information Processing Systems 34: Annual Conference on Neural Information Processing Systems 2021, NeurIPS 2021, December 6-14, 2021, virtual}, pages 9694--9705.

\bibitem[{Li et~al.(2019)Li, Yatskar, Yin, Hsieh, and Chang}]{li2019visualbert}
Liunian~Harold Li, Mark Yatskar, Da~Yin, Cho-Jui Hsieh, and Kai-Wei Chang. 2019.
\newblock Visualbert: A simple and performant baseline for vision and language.
\newblock \emph{arXiv preprint arXiv:1908.03557}.

\bibitem[{Li et~al.(2022{\natexlab{b}})Li, Liang, Zhao, Cui, Ouyang, Shao, Yu, and Yan}]{LiLZCOSYY22}
Yangguang Li, Feng Liang, Lichen Zhao, Yufeng Cui, Wanli Ouyang, Jing Shao, Fengwei Yu, and Junjie Yan. 2022{\natexlab{b}}.
\newblock \href {https://openreview.net/forum?id=zq1iJkNk3uN} {Supervision exists everywhere: {A} data efficient contrastive language-image pre-training paradigm}.
\newblock In \emph{Proc. of ICLR}. OpenReview.net.

\bibitem[{Lin et~al.(2014)Lin, Maire, Belongie, Hays, Perona, Ramanan, Doll{\'{a}}r, and Zitnick}]{DBLP:conf/eccv/LinMBHPRDZ14}
Tsung{-}Yi Lin, Michael Maire, Serge~J. Belongie, James Hays, Pietro Perona, Deva Ramanan, Piotr Doll{\'{a}}r, and C.~Lawrence Zitnick. 2014.
\newblock Microsoft {COCO:} common objects in context.
\newblock In \emph{Proc. of ECCV}.

\bibitem[{Maji et~al.(2013)Maji, Kannala, Rahtu, Blaschko, and Vedaldi}]{maji13fine-grained}
S.~Maji, J.~Kannala, E.~Rahtu, M.~Blaschko, and A.~Vedaldi. 2013.
\newblock \href {https://arxiv.org/abs/1306.5151} {Fine-grained visual classification of aircraft}.
\newblock Technical report.

\bibitem[{Mu et~al.(2022)Mu, Kirillov, Wagner, and Xie}]{MuK0X22}
Norman Mu, Alexander Kirillov, David~A. Wagner, and Saining Xie. 2022.
\newblock {SLIP:} self-supervision meets language-image pre-training.
\newblock In \emph{Proc. of ECCV}.

\bibitem[{Nilsback and Zisserman(2008)}]{nilsback2008automated}
M-E. Nilsback and A.~Zisserman. 2008.
\newblock Automated flower classification over a large number of classes.
\newblock In \emph{Proceedings of the Indian Conference on Computer Vision, Graphics and Image Processing}.

\bibitem[{Norelli et~al.(2022)Norelli, Fumero, Maiorca, Moschella, Rodol{\`a}, and Locatello}]{norelli2022asif}
Antonio Norelli, Marco Fumero, Valentino Maiorca, Luca Moschella, Emanuele Rodol{\`a}, and Francesco Locatello. 2022.
\newblock Asif: Coupled data turns unimodal models to multimodal without training.
\newblock \emph{ArXiv preprint}.

\bibitem[{Pedregosa et~al.(2011)Pedregosa, Varoquaux, Gramfort, Michel, Thirion, Grisel, Blondel, Prettenhofer, Weiss, Dubourg et~al.}]{pedregosa2011scikit}
Fabian Pedregosa, Ga{\"e}l Varoquaux, Alexandre Gramfort, Vincent Michel, Bertrand Thirion, Olivier Grisel, Mathieu Blondel, Peter Prettenhofer, Ron Weiss, Vincent Dubourg, et~al. 2011.
\newblock Scikit-learn: Machine learning in python.
\newblock \emph{the Journal of machine Learning research}.

\bibitem[{Plummer et~al.(2015)Plummer, Wang, Cervantes, Caicedo, Hockenmaier, and Lazebnik}]{DBLP:conf/iccv/PlummerWCCHL15}
Bryan~A. Plummer, Liwei Wang, Chris~M. Cervantes, Juan~C. Caicedo, Julia Hockenmaier, and Svetlana Lazebnik. 2015.
\newblock \href {https://doi.org/10.1109/ICCV.2015.303} {Flickr30k entities: Collecting region-to-phrase correspondences for richer image-to-sentence models}.
\newblock In \emph{2015 {IEEE} International Conference on Computer Vision, {ICCV} 2015, Santiago, Chile, December 7-13, 2015}, pages 2641--2649. {IEEE} Computer Society.

\bibitem[{Radenovic et~al.(2023)Radenovic, Dubey, Kadian, Mihaylov, Vandenhende, Patel, Wen, Ramanathan, and Mahajan}]{DiHT}
Filip Radenovic, Abhimanyu Dubey, Abhishek Kadian, Todor Mihaylov, Simon Vandenhende, Yash Patel, Yi~Wen, Vignesh Ramanathan, and Dhruv Mahajan. 2023.
\newblock Filtering, distillation, and hard negatives for vision-language pre-training.
\newblock \emph{CoRR}.

\bibitem[{Radford et~al.(2021)Radford, Kim, Hallacy, Ramesh, Goh, Agarwal, Sastry, Askell, Mishkin, Clark, Krueger, and Sutskever}]{RadfordKHRGASAM21}
Alec Radford, Jong~Wook Kim, Chris Hallacy, Aditya Ramesh, Gabriel Goh, Sandhini Agarwal, Girish Sastry, Amanda Askell, Pamela Mishkin, Jack Clark, Gretchen Krueger, and Ilya Sutskever. 2021.
\newblock \href {http://proceedings.mlr.press/v139/radford21a.html} {Learning transferable visual models from natural language supervision}.
\newblock In \emph{Proc. of ICML}, volume 139 of \emph{Proceedings of Machine Learning Research}, pages 8748--8763. {PMLR}.

\bibitem[{Raffel et~al.(2020)Raffel, Shazeer, Roberts, Lee, Narang, Matena, Zhou, Li, and Liu}]{raffel2020exploring}
Colin Raffel, Noam Shazeer, Adam Roberts, Katherine Lee, Sharan Narang, Michael Matena, Yanqi Zhou, Wei Li, and Peter~J. Liu. 2020.
\newblock \href {http://jmlr.org/papers/v21/20-074.html} {Exploring the limits of transfer learning with a unified text-to-text transformer}.
\newblock \emph{J. Mach. Learn. Res.}, 21:140:1--140:67.

\bibitem[{Reimers and Gurevych(2019)}]{reimers2019sentence}
Nils Reimers and Iryna Gurevych. 2019.
\newblock \href {https://doi.org/10.18653/v1/D19-1410} {Sentence-{BERT}: Sentence embeddings using {S}iamese {BERT}-networks}.
\newblock In \emph{Proc. of EMNLP}, pages 3982--3992, Hong Kong, China. Association for Computational Linguistics.

\bibitem[{Robinson et~al.(2021)Robinson, Chuang, Sra, and Jegelka}]{RobinsonCSJ21}
Joshua~David Robinson, Ching{-}Yao Chuang, Suvrit Sra, and Stefanie Jegelka. 2021.
\newblock \href {https://openreview.net/forum?id=CR1XOQ0UTh-} {Contrastive learning with hard negative samples}.
\newblock In \emph{Proc. of ICLR}. OpenReview.net.

\bibitem[{Schuhmann et~al.(2022)Schuhmann, Beaumont, Vencu, Gordon, Wightman, Cherti, Coombes, Katta, Mullis, Wortsman et~al.}]{schuhmann2022laion}
Christoph Schuhmann, Romain Beaumont, Richard Vencu, Cade Gordon, Ross Wightman, Mehdi Cherti, Theo Coombes, Aarush Katta, Clayton Mullis, Mitchell Wortsman, et~al. 2022.
\newblock Laion-5b: An open large-scale dataset for training next generation image-text models.
\newblock \emph{Advances in Neural Information Processing Systems}, 35:25278--25294.

\bibitem[{Shah et~al.(2022)Shah, Sra, Chellappa, and Cherian}]{ShahSCC22}
Anshul Shah, Suvrit Sra, Rama Chellappa, and Anoop Cherian. 2022.
\newblock \href {https://ojs.aaai.org/index.php/AAAI/article/view/20796} {Max-margin contrastive learning}.
\newblock In \emph{Thirty-Sixth {AAAI} Conference on Artificial Intelligence, {AAAI} 2022, Thirty-Fourth Conference on Innovative Applications of Artificial Intelligence, {IAAI} 2022, The Twelveth Symposium on Educational Advances in Artificial Intelligence, {EAAI} 2022 Virtual Event, February 22 - March 1, 2022}, pages 8220--8230. {AAAI} Press.

\bibitem[{Sharma et~al.(2018)Sharma, Ding, Goodman, and Soricut}]{SoricutDSG18}
Piyush Sharma, Nan Ding, Sebastian Goodman, and Radu Soricut. 2018.
\newblock \href {https://doi.org/10.18653/v1/P18-1238} {Conceptual captions: A cleaned, hypernymed, image alt-text dataset for automatic image captioning}.
\newblock In \emph{Proc. of ACL}, pages 2556--2565, Melbourne, Australia. Association for Computational Linguistics.

\bibitem[{Singh et~al.(2022)Singh, Gustafson, Adcock, de~Freitas~Reis, Gedik, Kosaraju, Mahajan, Girshick, Doll{\'a}r, and Van Der~Maaten}]{singh2022revisiting}
Mannat Singh, Laura Gustafson, Aaron Adcock, Vinicius de~Freitas~Reis, Bugra Gedik, Raj~Prateek Kosaraju, Dhruv Mahajan, Ross Girshick, Piotr Doll{\'a}r, and Laurens Van Der~Maaten. 2022.
\newblock Revisiting weakly supervised pre-training of visual perception models.
\newblock In \emph{Proceedings of the IEEE/CVF Conference on Computer Vision and Pattern Recognition}, pages 804--814.

\bibitem[{Stephenson et~al.(2021)Stephenson, Padhy, Ganesh, Hui, Tang, and Chung}]{stephenson2021geometry}
Cory Stephenson, Suchismita Padhy, Abhinav Ganesh, Yue Hui, Hanlin Tang, and SueYeon Chung. 2021.
\newblock \href {https://openreview.net/forum?id=V8jrrnwGbuc} {On the geometry of generalization and memorization in deep neural networks}.
\newblock In \emph{Proc. of ICLR}. OpenReview.net.

\bibitem[{Wah et~al.(2011)Wah, Branson, Welinder, Perona, and Belongie}]{wah2011caltech}
Catherine Wah, Steve Branson, Peter Welinder, Pietro Perona, and Serge Belongie. 2011.
\newblock The caltech-ucsd birds-200-2011 dataset.
\newblock Technical report, California Institute of Technology.

\bibitem[{Wang and Isola(2020)}]{wang2020understanding}
Tongzhou Wang and Phillip Isola. 2020.
\newblock \href {http://proceedings.mlr.press/v119/wang20k.html} {Understanding contrastive representation learning through alignment and uniformity on the hypersphere}.
\newblock In \emph{Proc. of ICML}, volume 119 of \emph{Proceedings of Machine Learning Research}, pages 9929--9939. {PMLR}.

\bibitem[{Wu et~al.(2022)Wu, Cheng, Zhang, Gao, Gonzalez, and Vajda}]{WuCZGGV22}
Bichen Wu, Ruizhe Cheng, Peizhao Zhang, Tianren Gao, Joseph~E. Gonzalez, and Peter Vajda. 2022.
\newblock \href {https://openreview.net/forum?id=G89-1yZLFHk} {Data efficient language-supervised zero-shot recognition with optimal transport distillation}.
\newblock In \emph{Proc. of ICLR}. OpenReview.net.

\bibitem[{Xiao et~al.(2010)Xiao, Hays, Ehinger, Oliva, and Torralba}]{xiao2010sun}
Jianxiong Xiao, James Hays, Krista~A. Ehinger, Aude Oliva, and Antonio Torralba. 2010.
\newblock \href {https://doi.org/10.1109/CVPR.2010.5539970} {{SUN} database: Large-scale scene recognition from abbey to zoo}.
\newblock In \emph{The Twenty-Third {IEEE} Conference on Computer Vision and Pattern Recognition, {CVPR} 2010, San Francisco, CA, USA, 13-18 June 2010}, pages 3485--3492. {IEEE} Computer Society.

\bibitem[{Yao et~al.(2022)Yao, Huang, Hou, Lu, Niu, Xu, Liang, Li, Jiang, and Xu}]{YaoHHLNXLLJX22}
Lewei Yao, Runhui Huang, Lu~Hou, Guansong Lu, Minzhe Niu, Hang Xu, Xiaodan Liang, Zhenguo Li, Xin Jiang, and Chunjing Xu. 2022.
\newblock \href {https://openreview.net/forum?id=cpDhcsEDC2} {{FILIP:} fine-grained interactive language-image pre-training}.
\newblock In \emph{Proc. of ICLR}. OpenReview.net.

\bibitem[{Zhang et~al.(2021)Zhang, Bengio, Hardt, Recht, and Vinyals}]{zhang2021understanding}
Chiyuan Zhang, Samy Bengio, Moritz Hardt, Benjamin Recht, and Oriol Vinyals. 2021.
\newblock Understanding deep learning (still) requires rethinking generalization.
\newblock \emph{Communications of the ACM}, 64(3):107--115.

\bibitem[{Zhang et~al.(2020)Zhang, Jiang, Wang, Kuang, Zhao, Zhu, Yu, Yang, and Wu}]{zhang2020devlbert}
Shengyu Zhang, Tan Jiang, Tan Wang, Kun Kuang, Zhou Zhao, Jianke Zhu, Jin Yu, Hongxia Yang, and Fei Wu. 2020.
\newblock Devlbert: Learning deconfounded visio-linguistic representations.
\newblock In \emph{Proceedings of the 28th ACM International Conference on Multimedia}, pages 4373--4382.

\end{thebibliography}

\appendix
\section{Appendix: Algorithm}
\label{apd:algs}
We summarize the Hard Pair Mining (HPM), the fast Hard Pair Mining (fastHPM) and the training pipeline of \hclip\ in Algorithm ~\ref{alg:hpm}, ~\ref{alg:fasthpm} and ~\ref{alg:finetuning} respectively.

\begin{algorithm}[h]
\DontPrintSemicolon
\caption{Hard Pair Mining (HPM)}
\label{alg:hpm}
\KwIn{Hard pairs number per sample $k$\;
Pretrained unimodal vision model: $f_{\text{text}}$\;
Pretrained unimodal vision model: $f_{\text{image}}$\;
Dataset $\mathcal{D} = \{ (x_1^{I}, x_1^{T}), (x_2^{I}, x_2^{T}), \cdots, (x_N^{I}, x_N^{T}) \}$\;
Threshold for visual and textual modality $\tau_{I}$\ and $\tau_{T}$\;
}
\KwOut{Hard samples $\mathcal{H} = [\mathcal{H}_1, \mathcal{H}_2, \cdots, \mathcal{H}_N]$}
\SetKwBlock{Begin}{function}{end function}{
\For{$i \in [1,N]$}{
    $\mathbf{s} \gets [0,0,\cdots, 0]^{\top} \in \mathbb{R}^{N}$\;
    $I_i \gets f_{\text{image}}(x_i^{I})$\;
    $T_i \gets f_{\text{text}}(x_i^{T})$\;
    \For{$j \in [1,N]$}{
        $I_j \gets f_{\text{image}}(x_j^{I})$\;
        $T_j \gets f_{\text{text}}(x_j^{T})$\;
        $\vec{S}^{I}_j \gets \frac{I_i \cdot I_j}{\left\|I_i\right\|_2 \cdot\left\|I_j\right\|_2} \, \textbf{if} \,\frac{I_i \cdot I_j}{\left\|I_i\right\|_2 \cdot\left\|I_j\right\|_2} > \tau_{I} \, \textbf{else} \, 0$\;
        $\vec{S}^{T}_j \gets \frac{T_i \cdot T_j}{\left\|I_i\right\|_2 \cdot\left\|T_j\right\|_2} \, \textbf{if} \,\frac{T_i \cdot T_j}{\left\|T_i\right\|_2 \cdot\left\|T_j\right\|_2} > \tau_{T} \, \textbf{else} \, 0$\;
        $\mathbf{s}_{j} \gets \vec{S}^{I}_j  \cdot \vec{S}^{T}_j$
    }
    $\mathcal{H}_i \gets \argmax(\mathbf{s}, k)$\;
      \uIf{$\exists j \in \mathcal{H}_i$, $\mathbf{s}_{j}=0$}{
        $\mathcal{H}_i = \emptyset$ \quad \#\, Indicate noise sample \;
        }
}}
\end{algorithm}

Note, in the inner for loop, shown in Algorithm~\ref{alg:hpm}, the image and caption representations will be repeatedly computed. To accelerate the hard pair mining and avoid unnecessary computational overhead, we compute and save the encoded image features and text features. Besides, the outer loop is parallelized in the implementation.

\begin{algorithm}[h]
\DontPrintSemicolon
\caption{fast Hard Pair Mining (fastHPM)}
\label{alg:fasthpm}
\KwIn{Hard pairs number per sample $k$\;
Pretrained unimodal vision model: $f_{\text{text}}$\;
Pretrained unimodal vision model: $f_{\text{image}}$\;
Dataset $\mathcal{D} = \{ (x_1^{I}, x_1^{T}), (x_2^{I}, x_2^{T}), \cdots, (x_N^{I}, x_N^{T}) \}$\;
Threshold for visual and textual modality $\tau_{I}$\ and $\tau_{T}$\;
\textcolor{orange}{Candidate pool size $C$}\;
}
\KwOut{Hard samples $\mathcal{H} = [\mathcal{H}_1, \mathcal{H}_2, \cdots, \mathcal{H}_N]$}
\SetKwBlock{Begin}{function}{end function}{
\For{$i \in [1,N]$}{
    \textcolor{orange}{Uniformly $C$ samples from Dataset $\mathcal{D}$, $\overline{\mathcal{D}}_i = \{ (x_1^{I}, x_1^{T}), (x_2^{I}, x_2^{T}), \cdots, (x_C^{I}, x_C^{T}) \}$}\;
    $\mathbf{s} \gets [0,0,\cdots, 0]^{\top} \in \mathbb{R}^{N}$\;
    $I_i \gets f_{\text{image}}(x_i^{I})$\;
    $T_i \gets f_{\text{text}}(x_i^{T})$\;
    \For{$j \in [1,C]$}{
        $I_j \gets f_{\text{image}}(x_j^{I})$\;
        $T_j \gets f_{\text{text}}(x_j^{T})$\;
        $\vec{S}^{I}_j \gets \frac{I_i \cdot I_j}{\left\|I_i\right\|_2 \cdot\left\|I_j\right\|_2} \, \textbf{if} \,\frac{I_i \cdot I_j}{\left\|I_i\right\|_2 \cdot\left\|I_j\right\|_2} > \tau_{I} \, \textbf{else} \, 0$\;
        $\vec{S}^{T}_j \gets \frac{T_i \cdot T_j}{\left\|I_i\right\|_2 \cdot\left\|T_j\right\|_2} \, \textbf{if} \,\frac{T_i \cdot T_j}{\left\|T_i\right\|_2 \cdot\left\|T_j\right\|_2} > \tau_{T} \, \textbf{else} \, 0$\;
        $\mathbf{s}_{j} \gets \vec{S}^{I}_j  \cdot \vec{S}^{T}_j$
    }
    $\mathcal{H}_i \gets \argmax(\mathbf{s}, k)$\;
      \uIf{$\exists j \in \mathcal{H}_i$, $\mathbf{s}_{j}=0$}{
        $\mathcal{H}_i = \emptyset$ \quad \#\, Indicate noise sample \;
        }
}}
\end{algorithm}

\begin{algorithm*}[h]
\caption{{H}ard sampl{E} for boosting contrastive {L}anguage-{I}mage {P}retrained models
(\hclip)}
\DontPrintSemicolon
\label{alg:finetuning}
    \KwIn{ $\mathcal{D} = \{ (x_1^{I}, x_1^{T}), (x_1^{I}, x_1^{T}), \cdots, (x_N^{I}, x_N^{T}) \}$\;
    Hard Pair Mining algorithm, $\text{HPM}()$ \quad \# or the $\text{fastHPM}()$  \;
    Pretrained unimodal vision model: $f_{\text{text}}$\;
    Pretrained unimodal vision model: $f_{\text{image}}$\;
    Pretrained contrastive language-image model $\{ \phi_{\text{image}}, \phi_{\text{text}} \}$ \;
    hyperparameters:\;
    \quad Hard pairs number $k$\;
    \quad Hard negative margin strength $\gamma$\;
    \quad Sampled hard negatives number $p$\;
    \quad Learning ratio $\eta$ \;
    \quad Batch size $b$ \;
    \quad Training iteration number $E$\;
    \quad Visual and textual modality threshold $\tau_{I}$\ and $\tau_{T}$\;
    
}
\KwOut{CLIP model $\{ \phi_{\text{image}}, \phi_{\text{text}} \}$} \;
$\mathcal{H} \gets \text{HPM}(\mathcal{D}, f_{\text{text}}, f_{\text{image}}, k, \tau_{I}, \tau_{T})$\;
\For {$iter \in [1,E]$}{
 $B \gets \{z_1, \ldots, z_b\} \stackrel{\text { i.i.d. }}{\sim} Uniform(\mathcal{D})$\;
    \For {$z_i \in B$}{
     $\mathcal{H}_{i}^{p} \gets \{z_i,  \ldots, z_p\} \stackrel{\text { i.i.d. }}{\sim} Uniform(\mathcal{H}_i)$ \;
     $\overline{B} \gets B\cup\mathcal{H}_{i}^{p}$\;
    }
 $\text{Compute loss} ~\ell_{\text{finetune}} \text{, Equation (6), with samples} ~\overline{B}$
 $\phi_{\text{image}} \gets \phi_{\text{image}} + \eta \cdot \partial_{\phi_{\text{image}}} \ell_{\text{finetune}}$
 $\phi_{\text{text}} \gets \phi_{\text{text}} + \eta \cdot \partial_{\phi_{\text{text}}} \ell_{\text{finetune}}$
}
\end{algorithm*}

\section{Appendix: Discussion about baselines}
\label{apd:baseline}
In our experiments, we utilized CLIP, SLIP, and DECLIP as baseline models on CC3M, CC12M, YFCC15M, and Open29M datasets. To ensure our results are both compelling and reproducible, we primarily employed publicly available checkpoints as our baseline and rigorously tested the effectiveness of \hclip\ against these checkpoints.
On CC3M, the checkpoint of SLIP model is released\footnote{https://github.com/facebookresearch/SLIP\#results-and-pre-trained-models}. 
We enhanced its performance by applying \hclip\, which notably improved the zero-shot performance on ImageNet from 23.00 to 26.05. However, we noticed that the CLIP with ResNet50 on CC3M is missing. To address this, we undertook the pretraining ourselves. Our results were encouraging: the performance of our pretrained CLIP with ResNet50 achieved a score of 19.86, surpassing the 17.10 achieved by SLIP's CLIP with ViT-B/32 as reported in~\cite{MuK0X22}. This outcome suggests the robustness of our implementation. Besides, consistent with several prior studies, we found that on smaller pretraining datasets, CLIP with ResNet50 outperforms CLIP with ViT-B.
On the CC12M dataset, a similar situation arose: while the SLIP checkpoint was available, the CLIP model was absent, leading us to undertake its pretraining. 
On the YFCC15M (v1) collected by ~\cite{RadfordKHRGASAM21}, we trained the CLIP model. This resulted in a 25.46 score in the ImageNet zero-shot classification, closely aligning with the 26.10 outcome reported by ~\cite{cui2022democratizing}. Additionally, for the YFCC15M (v2) dataset referenced in \cite{LiLZCOSYY22},  both SLIP and DECLIP pretrained parameters were made available by \cite{LiLZCOSYY22}, which we utilized directly as our baselines. 
On the larger dataset, Open29M, there was a lack of open-source pretrained checkpoints, prompting us to conduct the pretraining ourselves. Notably, the performance of our reimplementation (42.32) closely aligns with the results reported by \cite{LiLZCOSYY22}, indicating the effectiveness of our approach.

\section{Appendix: Analysis of the Impact of Subset Size on Hard Pair Selection in FastHPM}
\label{apd:subset_size}
\begin{figure*}[h]
\centering
\includegraphics[width=\linewidth]{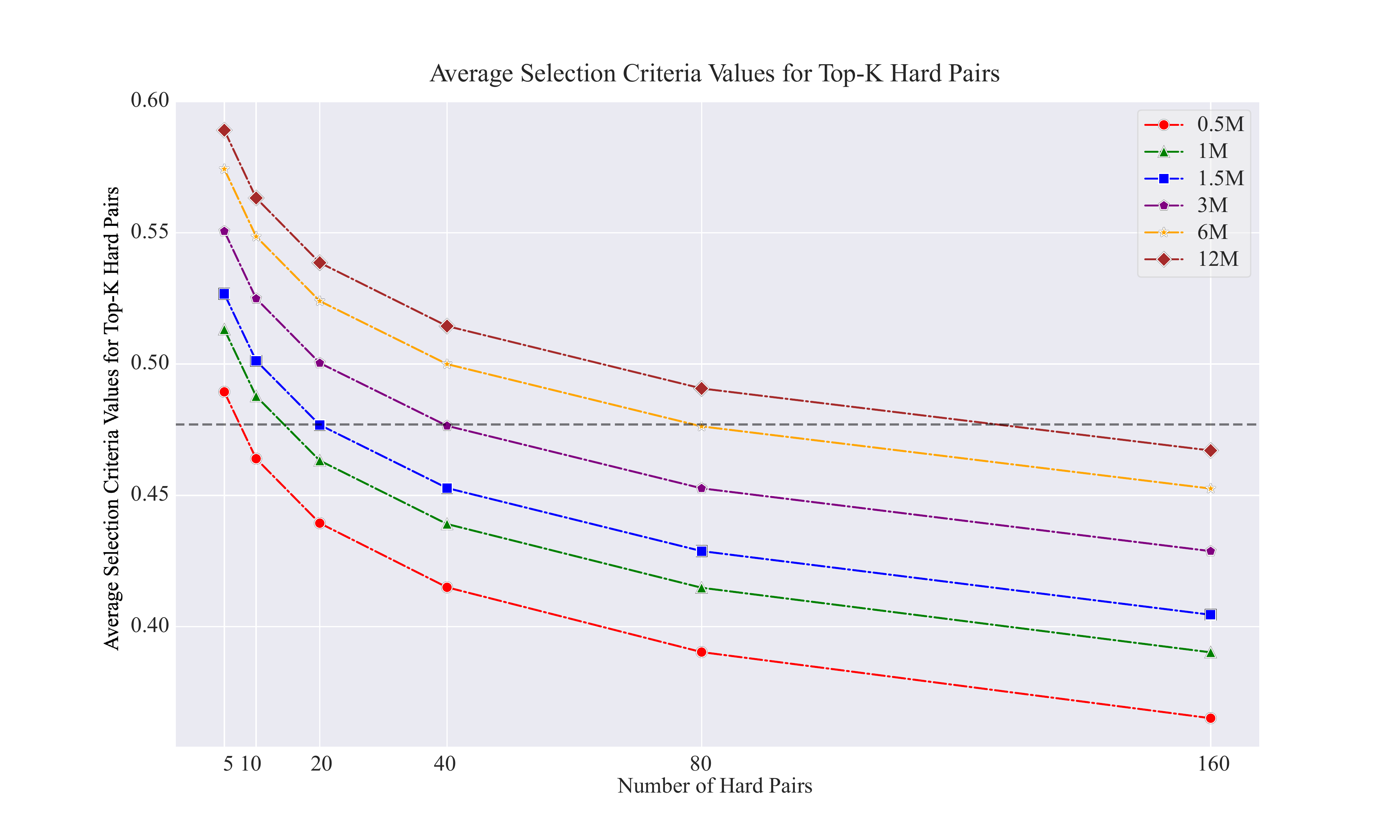}
\caption{The average selection criteria values for hard pairs mined by FastHPM with different subset sizes.
\label{fig:avg_selection_criteria}
}
\end{figure*}

In the comparison of HPM and FastHPM detailed in Section~\ref{subsec:hpm_detail}, we explore the efficacy of using 3M  and 6M subset sizes of the CC12M dataset in FastHPM for mining hard pairs.
The result, Table~\ref{tab:approx}, shows that with a reduced subset size as small as 3 million entries, mining hard pairs and further training with these pairs can boost CLIP to achieve competitive performance with full set for mining.

In this section, we delve deeper into the analysis of hard pairs mined by FastHPM across varying subset sizes. Based on the selection criteria defined by FastHPM (Equation~\ref{equ:sample_selection_approx}), we denote the \textit{selection criteria value} as $\widetilde{S}^{I}(x^I_i, \mathcal{H}^{\star}_i(j))^{\top} \widetilde{S}^{T}(x^T_i, \mathcal{H}^{\star}_i(j))$. Here, $\mathcal{H}^{\star}_i(\cdot)$ represents a pair within the set of hard pairs $\mathcal{H}^{\star}_i$, mined by FastHPM for a specified target pair $i$ under a given subset size.
Additionally, the $j$ in $\mathcal{H}^{\star}_i(j)$ indicates the $j$-th hard pair within the set $\mathcal{H}^{\star}_i$. Note, a higher selection criteria value signifies a harder mined pair.

We present the average selection criteria values for top-k hard pairs in Figure~\ref{fig:avg_selection_criteria}. As depicted by the grey horizontal line, the average selection criteria values for the top-20 hard pairs selected by FastHPM-1.5M, the top-40 by FastHPM-3M, and the top-80 by FastHPM-6M all approximate 0.477. 
This figure indicates that a further reduction in the subset size might necessitate adjustments to the number of hard pairs sampled to preserve quality. For instance, in our experiments detailed in Table~\ref{tab:approx}, we uniformly sampled hard pairs for training from the top 50 for \hclip-3M. As Figure~\ref{fig:avg_selection_criteria} suggests, a sampling range of 10 for \hclip-1M might be effective. Particularly, considering that \hclip significantly boosted the pre-trained models with just an additional training epoch, as discussed in Section~\ref{subsec:main_exp}, selecting one hard pair for each target pair from a pool of 10 will be feasible.

\section{Appendix: Implementation Details}
Our experiments are conducted across three distinct architectures: ResNet-50, ViT-B/16, and ViT-B/32, tailored to various datasets and pretrained models. Specifically, for loading the pretrained CLIP model on CC3M and CC12M, the ResNet-50 is used as the image encoder. Besides, to align with existing checkpoints established by ~\citet{MuK0X22}, we use ViT-B/16 for SLIP model experiments on CC3M and CC12M, respectively. And, we use ViT-B/32 for pretraining on YFCC15M v1, v2, and Open29M datasets to ensure fair comparison with the previous results~\cite{LiLZCOSYY22}. 
Furthermore, for the SLIP and DECLIP models, we adapt the pretrained parameters from the publicly available resources\footnote{\url{https://github.com/facebookresearch/SLIP}, \url{https://github.com/Sense-GVT/DeCLIP}.}
The input resolution of the image encoder is 224 $\times$ 224 and the maximum context length of the text encoder is 77. All of our experiments are conducted on 8 V100 GPUs with a batch size of 128 for ViT-B/16 models, and a batch size of 512 for ResNet-50 models and ViT-B/32 models. The dimension of the image and text embeddings is 1024 for ResNet-50 models and 512 for ViT-B/16 and ViT-B/32 models.
We set $\tau=0.5$, $\gamma=1$, $k=50$ and $p=1$ for all the experiments by default. Automatic mixed-precision is used to save GPU memory.
To keep the model from overfitting, we use early stopping if there is no performance gain on ImageNet zero-shot accuracy in 5 epochs.
It is worth noting that using zero-shot classification performance on ImageNet as a criterion for early stopping is a commonly used practice for the training of CLIP~\citep{RadfordKHRGASAM21, MuK0X22}.

To reflect that our method is designed to work with few assumptions on encoder, we used encoders pretrained over a single-modal source rather than multimodally pretrained ones when preparing hard negative pairs. 
Specifically, we used an unsupervised pre-trained vision transformer, DINO VITs8~\citep{caron2021emerging}, and a Sentence Transformer (SentenceT)~\citep{reimers2019sentence} to encode text. For DINO VITs8, the embedding size is 384, while for SentenceT, it is 768.

\section{Appendix: Performance of HELIP on noisy dataset}
\label{subsec:noisy_data}
We expanded our investigation to assess the effectiveness of \hclip on subsets of LAION7.5M and 8M, which are randomly sampled from LAION~\citep{schuhmann2022laion}. The results are detailed in Table~\ref{tab:laion6m}.
The CLIP model, enhanced with \hclip\, consistently outperformed its original counterpart on both subsets across a majority of the evaluated datasets, including ImageNet, CIFAR10, CIFAR100, Caltech, and Food.
On the 7.5M subset, \hclip\ enhances performance across all datasets by an average of 3.6\%. Although CLIP scores slightly higher on the Sun dataset, \hclip boosts its overall performance with an average improvement of 2.5\% on the 8M subset.
These results highlight the enhanced performance achieved through \hclip, demonstrating its robustness and effectiveness in improving existing models that have been pretrained on noisy data.
\begin{table*}[h]
\vspace{-5pt}
\centering
\small
\begin{tabular}{cccccccc}
\toprule
     & ImageNet & CIFAR10 & CIFAR100 &Caltech & Food & Sun &Avg. \\ 
\toprule
CLIP-7.5M &  23.5 &   34.6   & 14.5  & 58.9 & 28.6 & 25.3 &  30.8        \\
CLIP-\hclip-7.5M &  \textbf{25.8} &  \textbf{39.9} & \textbf{16.7} & \textbf{61.9} & \textbf{34.1} & \textbf{28.2}     & \textbf{34.4}        \\
CLIP-8M &  25.1 &  31.1 & 12.9  & 60.9 & 29.5 & \textbf{27.5} & 31.2        \\
CLIP-\hclip-8M &  \textbf{26.5} & \textbf{38.8} & \textbf{14.6} & \textbf{62.3} & \textbf{33.1} & 26.6 & \textbf{33.7}\\
\bottomrule
\end{tabular}
\caption{{Zero-shot performance of CLIP on two LAION subsets.}
\label{tab:laion6m}
} 
\end{table*}

\section{Appendix: Analysis of the Impact of $\tau$ on Hard Pair Selection}
\begin{figure}[h]
\includegraphics[width=1.1\linewidth]{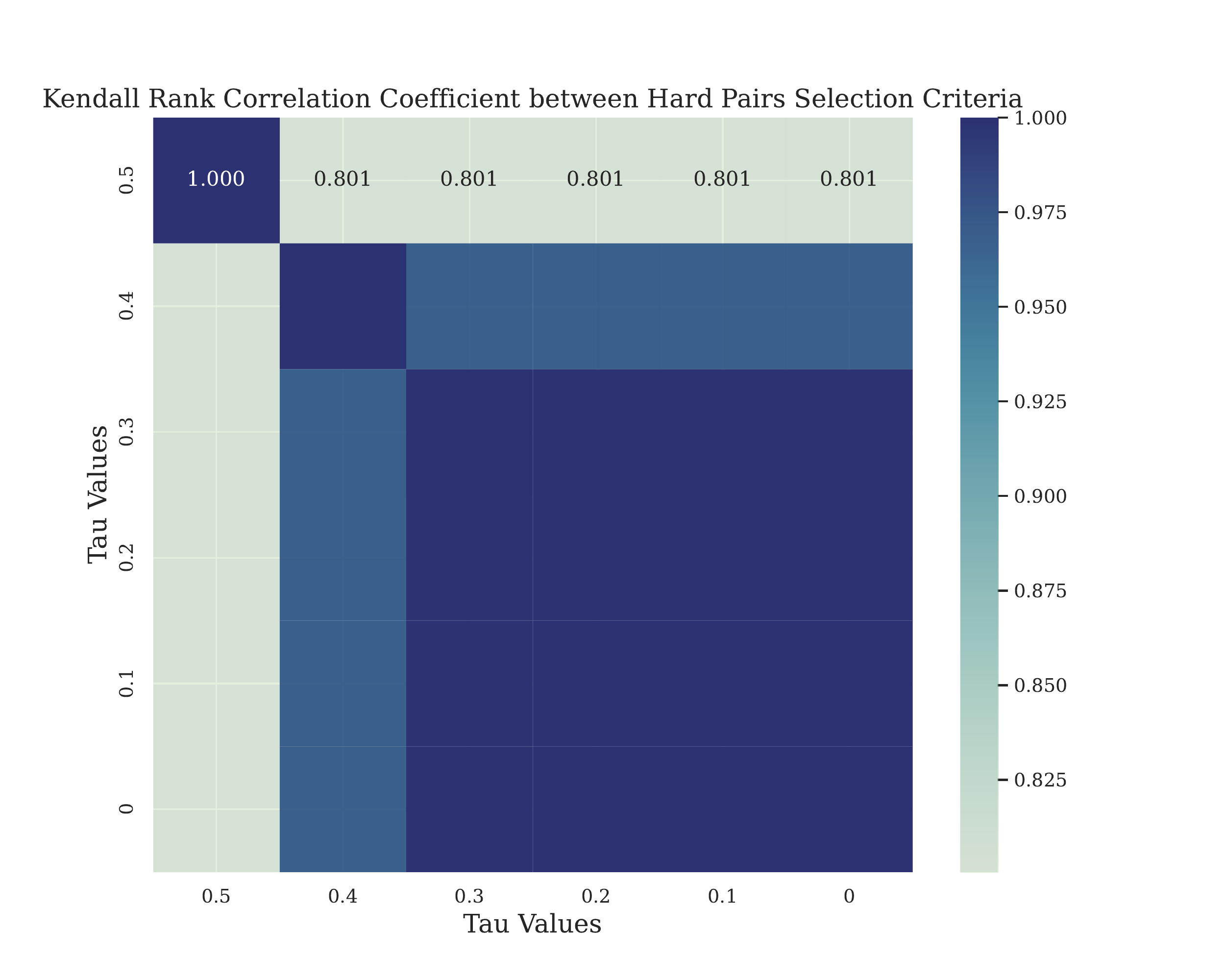}
\caption{The Impact of $\tau$ on Hard Pair Selection.\label{fig:tau_impact}
}
\end{figure}

To examine the impact of the threshold parameter $\tau$ on the selection of hard pairs, we analyze the similarities in the rankings of hard pairs (using Kendall Rank Similarity) mined by HPM under various $\tau$ values. 
The hard pairs are ranked by using the selection criteria value mentioned in Appendix~\ref{apd:subset_size}.
The results on the CC12M dataset are displayed in Figure~\ref{fig:tau_impact}.
We observe that the selection of hard pairs is robust to changes in the $\tau$ value. This resilience is partly because we only mine the top 50 hard pairs, a subset unlikely to be significantly affected when $\tau \leq 0.5$.

\section{Appendix: Analysis of the Impact of Mitigating Noisy Data}
As presented in Section~\ref{sec:hard_sample_mining}, to enhance the overall quality and reliability of the training dataset, data pairs lacking substantial support from the entirety of the training data are considered unsuitable and removed.

\begin{figure*}[h]
\includegraphics[width=\linewidth]{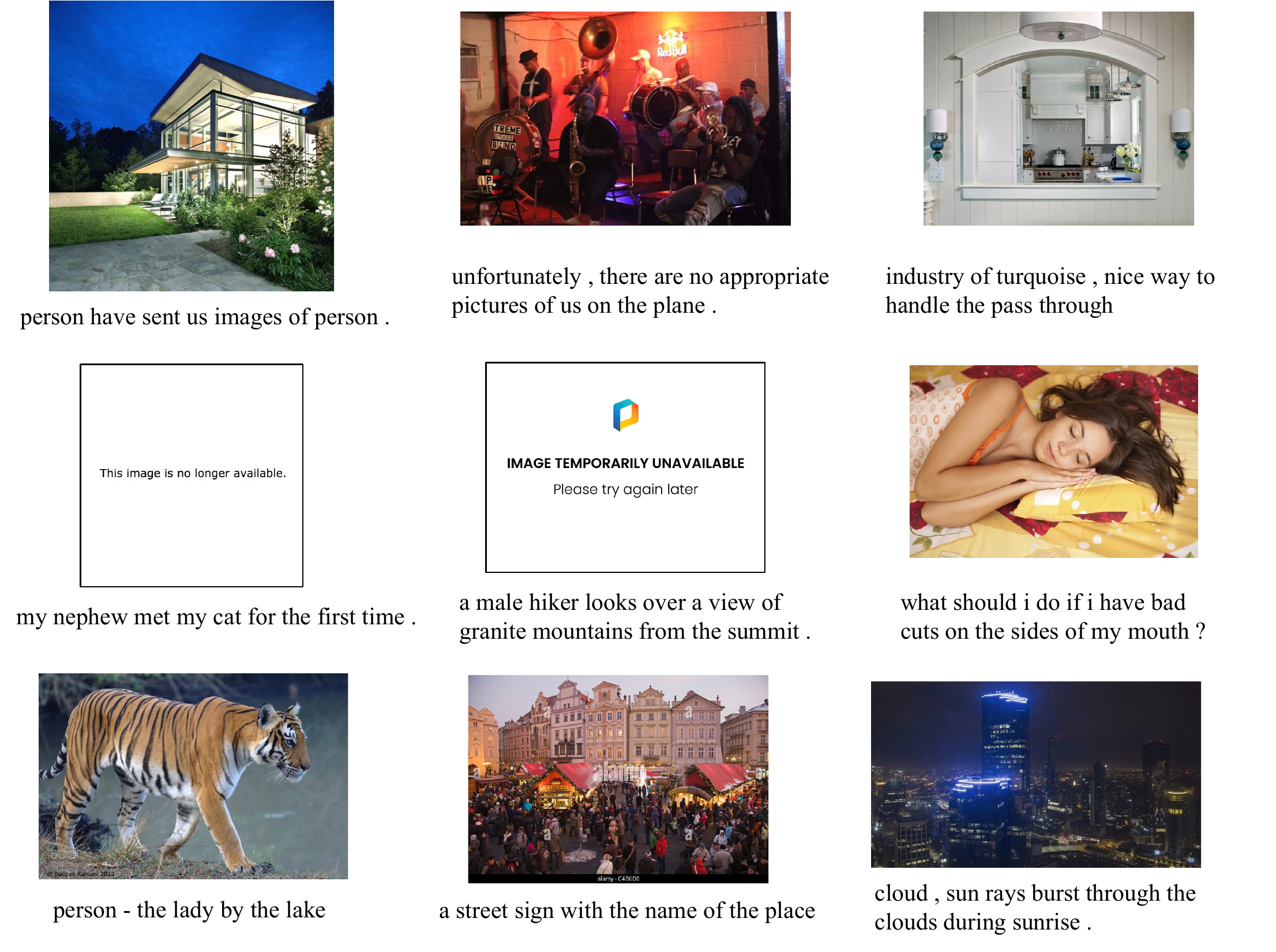}
\caption{Visualization of the image-caption pairs filtered out from CC12M.\label{fig:filter_out}}
\end{figure*}

This section further empirically analyzes the impact of our noise mitigation strategy by detailing the quantity and nature of pairs removed across various datasets. Specifically, our approach removes 4.67\% of the pairs from CC3M, 3.64\% from CC12M, and 7.41\% from YFCC15M, before continuing with pretraining. Figure~\ref{fig:filter_out} visualizes the pairs filtered from CC12M. Notably, our strategy effectively removed pairs such as unavailable images (e.g., two blank or white images in the second row) and mismatched pairs. These results suggest that our noise mitigation strategy can effectively clean the data using two single-modality models before training a CLIP model from scratch.

\section{Appendix: Comparison with Other Hard Data Selection Method}
\label{subsec:other_hard_data_selection}
\begin{figure}[h]
\centering
\includegraphics[width=\linewidth]{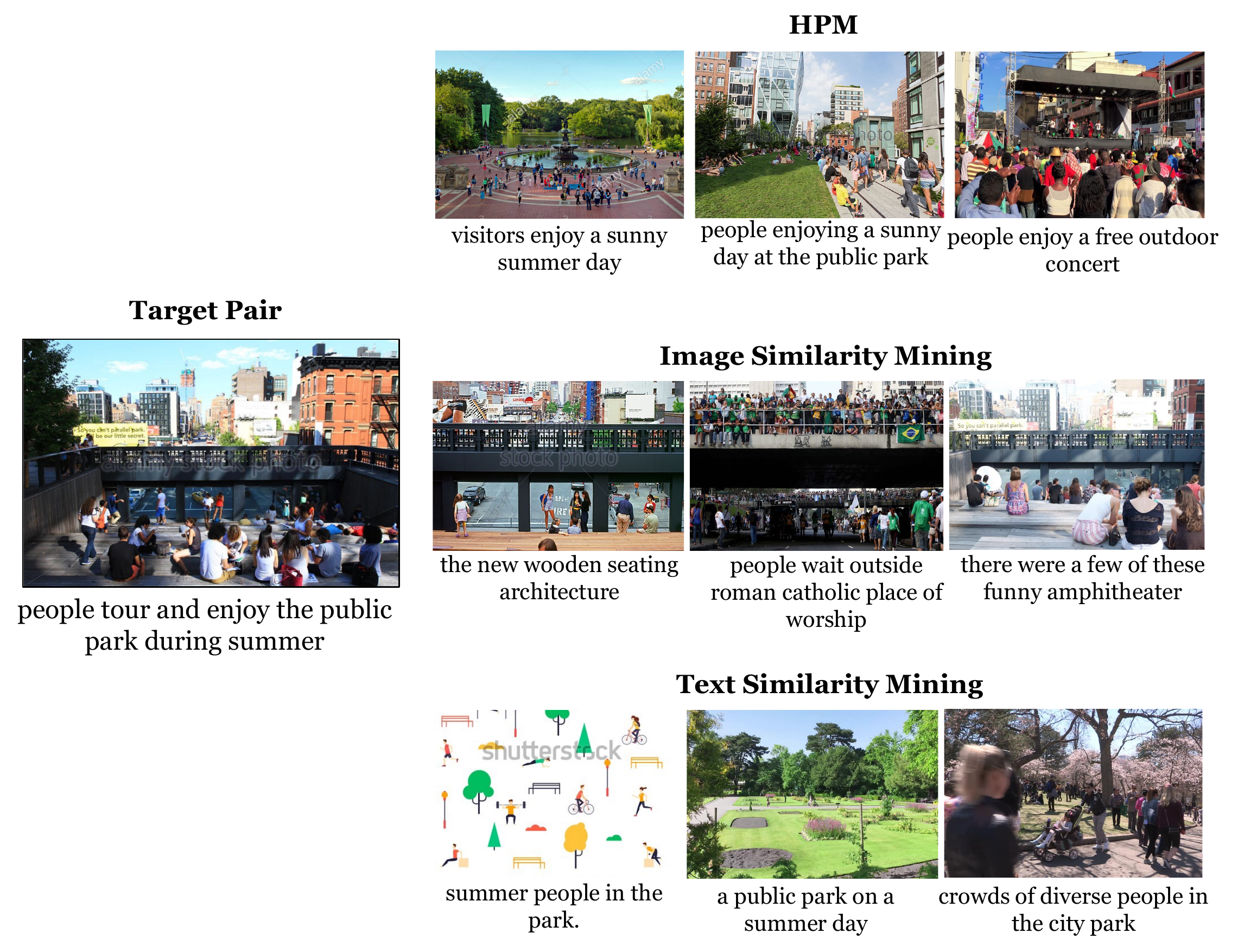}
\caption{\textbf{Hard negative data selected by different methods.} Compared to data mined using the sample-level (image/text modal) similarity, hard pairs mined by HPM are more similar to the target.} 
\vspace{-10pt}
\label{fig:mining_methods_vis}
\end{figure}

We evaluate the efficacy of the proposed method in enhancing the discriminative capacity of learned representations by comparing its zero-shot classification performance with that of other hard data mining strategies. As described in the Section~\ref{sec:Related Work}, a common way to define hard data is through intra-modality similarity. Hence, we introduce the hard data mining methods depending on (sample level) image similarity mining and text similarity mining and denote them as IM and TM respectively. 
For a given target pair, we compute the cosine similarity between its image/text representation and that of the remaining dataset. The image and text representations are encoded using a pretrained Resnet50 and BERT, respectively. As the preprocessing step, IM and TM methods mine hard negatives before continuous pretraining. Subsequently, we integrate the mined \tmlrnew{hard negative pairs} into the training pipeline of CLIP and denote them as CLIP+IM and CLIP+TM and optimize the original contrastive loss to fine-tune the model. Additionally, we also include the hard negative contrastive loss, HN-NCE, proposed by ~\citet{DiHT}, as a baseline. HN-NCE upsamples the weight of hard-negatives identified by the current model. As shown in Table~\ref{tab:hard_sample_methods}, when the CC3M pretrained CLIP model is combined with \hclip, the performance of our pair-level hard data mining method significantly outperforms other sample-level techniques. 
Besides,we observe that compared to the baseline CLIP performance, the introduction of TM and IM methods results in a decline in performance. To better understand the reasons behind this drop, we analyzed the outputs of the TM and IM methods in detail.
In Figure\ref{fig:mining_methods_vis}, we illustrate the data obtained through three distinct preprocessing methods: Hard Pair Mining (HPM), Image Similarity Mining (IM), and Text Similarity Mining (TM). The first row depicts the image-text pairs identified by HPM, while the second and third rows showcase the pairs mined by IM and TM, respectively.
For TM (IM displays similar issues), the selected pairs often feature captions that are highly similar or identical, which is typical in data collected from the web. Even though identical pairs may not always be present, repetitions of the same images or text are common. 
According to the CLIP contrastive loss (Equation~\ref{equ:clip_loss}), the model is forced to push nearly identical caption representations toward and away from two distinct image representations at the same time.This inherent contradiction in objectives contributes to a degradation in performance. 
To illustrate, consider a target pair \((T_{\text{target}}, I_{\text{target}})\) and a mined pair \((T_{\text{mined}}, I_{\text{mined}})\) using TM, where \( T_{\text{target}} \approx T_{\text{mined}} \) but \( I_{\text{target}} \not\approx I_{\text{mined}} \). In the contrastive loss framework, the model aims to minimize the distance between \((I_{\text{target}}, T_{\text{target}})\) and maximize the distance between \((I_{\text{target}}, T_{\text{mined}})\). However, the near-identity of \(T_{\text{target}}\) and \(T_{\text{mined}}\) leads to conflicting optimization targets and a potential decline in performance.

\begin{table}[t]
\centering
\small
\begin{tabular}{lccc}
\toprule
             & Imagenet & CIFAR10 & CIFAR100 \\
\toprule
CLIP & 19.04       & {33.06} & {13.77} \\
\midrule
CLIP + TM    &  16.70         &  28.71         &    9.67          \\
CLIP + IM   &     16.93       &   29.22        &  10.42               \\ 
CLIP + HN-NCE   &    19.47     &  29.88           &  11.83                   \\ 
CLIP + \hclip  &     \textbf{19.86}       &    \textbf{34.05}        &     \textbf{14.13}  \\
\bottomrule
\end{tabular}
\caption{
\textbf{Zero-shot performance of CLIP pre-trained on CC3M boosted by hard data mined by different methods.} \hclip\ shows superior performance, consistently outperforming local/global hard sample mining techniques by a substantial margin.
\label{tab:hard_sample_methods}
}
\vspace{-10pt}
\end{table}

\section{Appendix: Impact of different encoders in HPM}
\label{subsec:different_encoders}
We explored the effect of different pretrained encoders on HPM's performance by alternating image and text encoders. Initially, the unsupervised pretrained DINO VITs8~\citep{caron2021emerging} was paired with the SentenceT~\citep{reimers2019sentence} transformer, trained on over a billion internet-based sentences. This combination was later swapped for the SWAG VITb16~\citep{singh2022revisiting} and the T5~\citep{raffel2020exploring}. Additionally, experiments using OpenAI's CLIP model~\citep{RadfordKHRGASAM21} multimodal encoders were conducted. Interestingly, as Table~\ref{tab:encoders} suggests, the encoder choice seemingly has negligible impact on HPM's performance, likely due to the proficiency of current pretrained models in modeling intra-modal similarities. 
Moreover, the ability to use single-modal pretrained models and still achieve competitive or superior performance implies that there's no assumption of having access to a high-quality CLIP model, such as OpenAI's CLIP-400M.

\begin{table}[h]
    \centering
    \small
    \setlength{\tabcolsep}{2pt}{
    \begin{tabular}{ccccc}
    \toprule
         & ImageNet & CIFAR10 & CIFAR100 & Avg. \\ 
    \toprule
    {CLIP Encoders} &  {19.57}  &  {33.28}  &  {13.53} &   {22.12}      \\
    VITs8+SentenceT &  19.86 &   34.05   & 14.13  & 22.68        \\
    VITb16+SentenceT &  19.62  &  35.53  &  14.67 &   23.27    \\
    VITs8 + T5 &  19.61  &  33.99  &  13.82 &   22.47     \\
    \bottomrule
    \end{tabular}
    \caption{{The zero-shot performances of HELIP with different encoders in HPM.}
    HPM's performance is insensitive to the selection of encoders. \label{tab:encoders}
    }
    }
\end{table}

\section{Appendix: More visualization results}
\label{apd:more_exp}
\begin{figure*}[h]
\begin{center}
{\includegraphics[width=\linewidth]{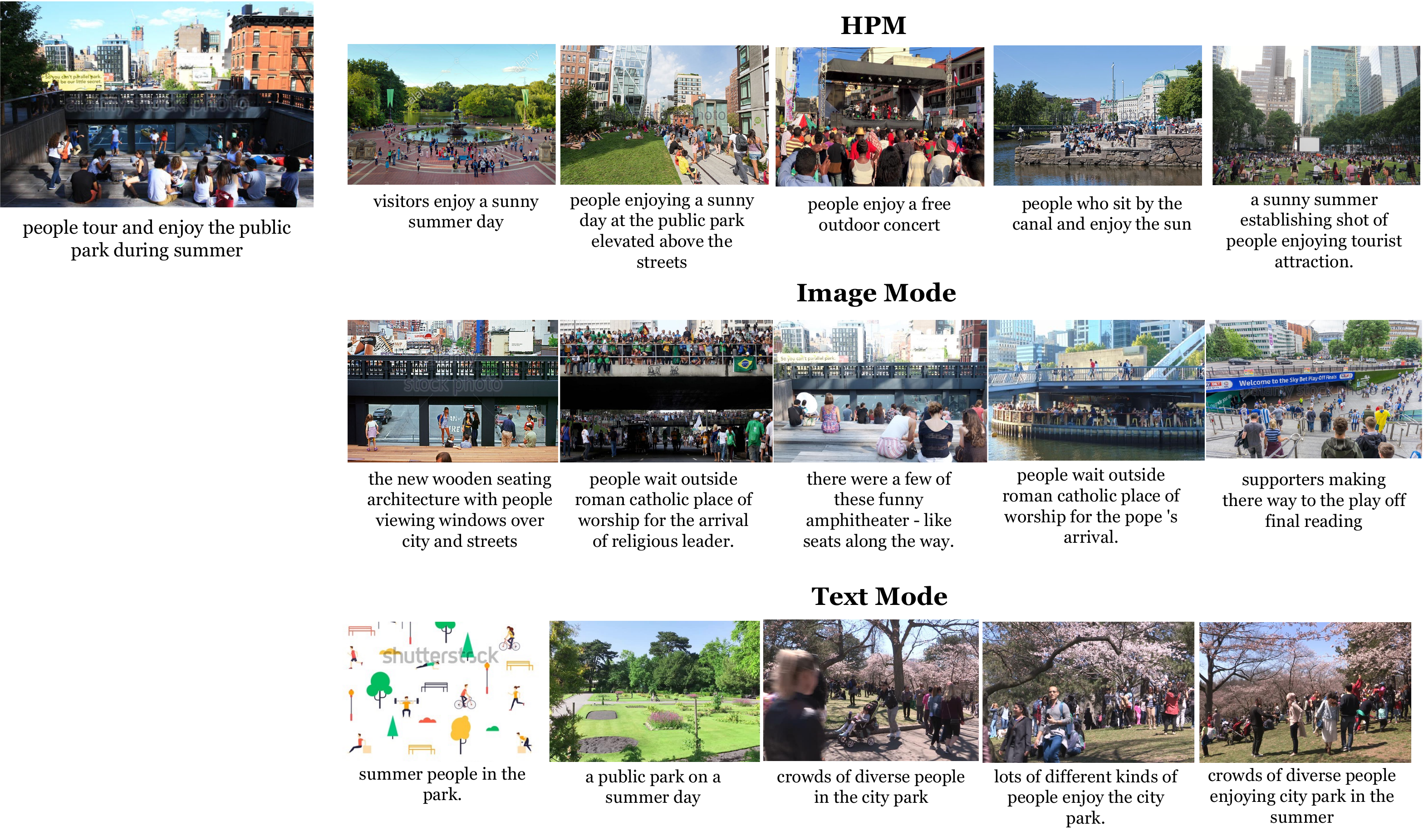}}
\hfill
{\includegraphics[width=\linewidth]{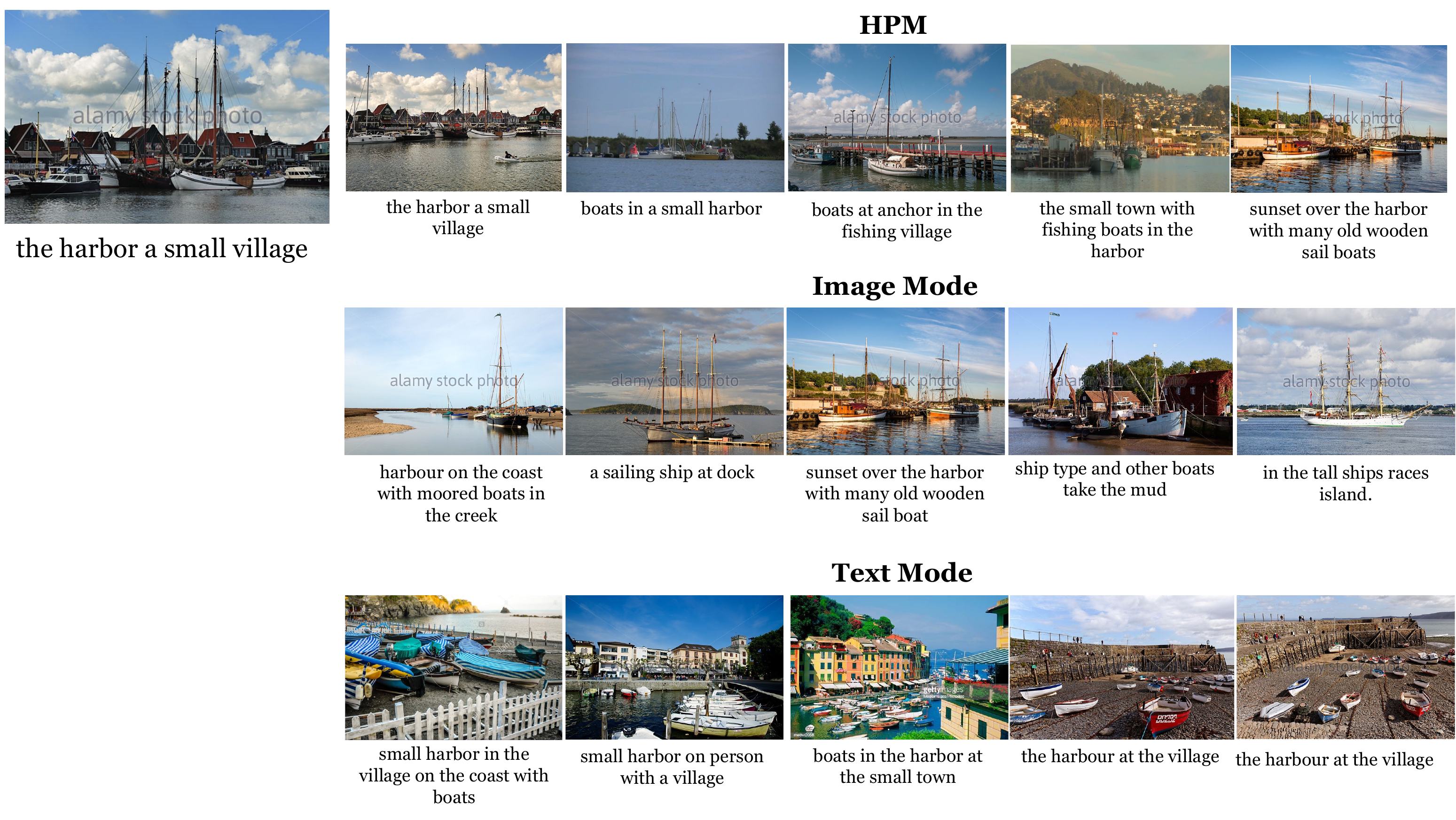}}
\end{center}
   \caption{\textbf{Hard pairs selected by different methods.\label{fig:more_minining_comparison}}  
   }
\end{figure*}

We offer further visualization results pertaining to the hard samples mined by various methods. As depicted in Figure~\ref{fig:more_minining_comparison}, the hard samples sourced by HPM closely resemble the target sample (seen at the top left). Interestingly, for samples with fewer objectives, the image and text mining method can identify a reasonably challenging counterpart, as seen in the case of ``the harbor in a small village''. However, for intricate scenes, only the HPM is capable of yielding sufficiently challenging samples, like the scenario ``people touring and enjoying the public park during summer''. The dataset acquired from the web encompasses a myriad of such intricate cases. We posit that this is why training with hard samples unearthed by HPM yields more proficient outcomes.

Moreover, we present additional visualization results for hard samples mined via different techniques. Hard samples extracted by HPM exhibit a stronger resemblance to the target sample, as highlighted in Figure~\ref{fig:more_minining_comparison} (top left). We observed that the image and text mining methods can provide a relatively fitting hard counterpart for simpler samples, like ``the harbor in a quaint settlement''. However, for more intricate scenes, only the HPM method produces samples of adequate difficulty, such as ``people touring and relishing the public park throughout summer''. The web-based dataset includes a significant proportion of these complex cases. Consequently, we infer that training with hard samples mined by HPM results in enhanced performance.

\section{Appendix: Future work}
\label{apd:future}
\vspace{-5pt}
Moving forward, several possibilities for future research emerge. First, we aim to explore composition-aware fine-tuning for VLMs, which could potentially enable more effective utilization of multimodal information. Moreover, we are intrigued by the prospect of combining parameter-efficient tuning~\citep{he2022towards} with \hclip\, potentially further enhancing performance. Another area of interest is scaling up the dataset size and examining the applicability of the scaling law to our method. We also intend to investigate how the integration of our boosting algorithm might alter the multimodal dataset curation algorithm~\citep{gadre2023datacomp}. Ultimately, we hope our work will serve as a catalyst for additional research in the fine-tuning of pre-trained, large-scale multimodal models.

\section{Appendix: Potential Risks and Ethical Considerations}
While our research contributes to advancements in language-image pretraining, it is important to acknowledge potential risks and ethical considerations associated with our work.

\noindent\textbf{Privacy Concerns:} The datasets used—CC3M, CC12M, subsets of YFCC15M, LAION7.5M, and LAION8M—are publicly available and sourced from the internet.
They may contain personal identifiable information (PII) or images of individuals.
We rely on the dataset providers' curation processes to remove personal identifiable information.

\noindent\textbf{Bias and Fairness Issues:} The datasets may not be representative of all demographics or cultures, potentially leading to models that perform unevenly across different groups. This lack of fairness can perpetuate existing societal biases and inequalities.

\noindent\textbf{Misuse of Technology:} The models developed could be misused for malicious purposes, such as generating deepfakes, enabling unauthorized surveillance, or creating misleading information, which could have negative societal impacts.\\

\noindent To address these potential risks, the following steps could be considered in future work:\\

\noindent\textbf{Enhanced Data Filtering:} 
Currently, we rely on the curation and filtering processes conducted by the dataset creators for the datasets we used—CC3M, CC12M, YFCC15M subsets, and subsets of LAION. Implementing additional data cleaning procedures to identify and remove personally identifiable information (PII) and offensive content from the training datasets is an important area deserving future study.

\noindent\textbf{Bias Mitigation Techniques:} Incorporate fairness-aware learning algorithms and conduct thorough evaluations to detect and reduce biases in the model's output.

\noindent\textbf{Transparency and Accountability:} Provide documentation detailing the data sources, model limitations, and potential biases to inform users and stakeholders.

\section{Appendix: Licensing}
All datasets used in this study are open-source and utilized in accordance with their respective licenses. Specifically, CC3M and CC12M comprise images with Creative Commons licenses, YFCC100M is distributed under a Creative Commons Attribution license, and LAION-5B consists of web-sourced data under permissive licenses. Our use of these datasets is strictly for research purposes and complies with their licensing terms.

\section{Appendix: Use of AI Assistants}
We acknowledge the use of AI assistants in the preparation of this work. Specifically, we utilized tools, GPT4-o, for proofreading the manuscript and GitHub Copilot for assisting with coding tasks. These AI tools were employed to enhance productivity and efficiency. All content generated with the assistance of AI was thoroughly reviewed and edited by the authors to ensure accuracy and originality. The responsibility for the final content of this paper rests solely with the authors.

\section{Appendix: Artifact Use Consistent with Intended Purpose}

In our research, we utilized several publicly available datasets—CC3M, CC12M, YFCC15M subsets, LAION7.5M, and LAION8M—that are widely used within the computer vision and machine learning communities for the purpose of training and evaluating language-image models. The use of these datasets in our study is consistent with their intended purpose, as specified by their creators, which is to advance research in image recognition, captioning, and related fields. All datasets were used strictly for non-commercial, research-oriented objectives, adhering to the access conditions and licenses provided by the dataset providers.

For the artifacts we created during this research, including the trained models and any derived datasets or code, we specify that their intended use is for academic and research purposes only. These artifacts are shared to promote transparency, reproducibility, and further advancement in the field. We ensure that this intended use is compatible with the original access conditions of the datasets we used, particularly considering any restrictions on derivative works or redistribution. When sharing our artifacts, we comply with all applicable licenses and access terms, and we encourage others who use our artifacts to do the same.

\end{document}